\documentclass[final]{cvpr}
\usepackage{nopageno}
\usepackage{times}
\usepackage{epsfig}
\usepackage{graphicx}
\usepackage{amsmath}
\usepackage{amssymb}

\usepackage[dvipsnames]{xcolor}
\usepackage{kotex}
\usepackage{multirow}
\usepackage{booktabs}
\usepackage{makecell}
\usepackage[export]{adjustbox}
\usepackage{upquote}
\usepackage{stackengine} 
\usepackage{scalerel}
\usepackage{soul}
\usepackage{tabularx}
\newcolumntype{Y}{>{\centering\arraybackslash}X}

\makeatletter
\@namedef{ver@everyshi.sty}{}
\makeatother
\usepackage{tikz}

\usepackage[pagebackref=true,breaklinks=true,colorlinks,bookmarks=false]{hyperref}
\usepackage{algorithm} 
\usepackage{algpseudocode} 
\usepackage{eqparbox}
\usepackage{colortbl}

\def\cvprPaperID{208} %
\def\confYear{CVPR 2021}

\begin{document}

\newcommand{\Eq}[1]  {Eq.\ (#1)}
\newcommand{\Eqs}[1] {Eqs.\ (#1)}
\newcommand{\Fig}[1] {Fig.\ #1}
\newcommand{\Figs}[1]{Figs.\ #1}
\newcommand{\Tbl}[1]  {Table\ #1}
\newcommand{\Tbls}[1] {Tables\ #1}
\newcommand{\Sec}[1] {Sec.\ #1}
\newcommand{\SSec}[1] {Sec.\ #1}
\newcommand{\Secs}[1] {Secs.\ #1}
\newcommand{\Alg}[1] {Alg.\ #1}
\newcommand{\Etal}   {{\textit{et al.}}}

\newcommand{\setone}[1] {\left\{ #1 \right\}} %
\newcommand{\settwo}[2] {\left\{ #1 \mid #2 \right\}} %

\newcommand{\todo}[1]{{\textcolor{red}{#1}}}
\newcommand{\son}[1]{{\textcolor{magenta}{hyeongseok: #1}}}
\newcommand{\jy}[1]{{\textbf{\textcolor{MidnightBlue}{[JY] }}\textcolor{MidnightBlue}{#1}}}
\newcommand{\sean}[1]{{\textcolor{green}{sean: #1}}}
\newcommand{\sunghyun}[1]{{\textcolor[rgb]{0.6,0.0,0.6}{sunghyun: #1}}}
\newcommand{\jsrim}[1]{{\textcolor[rgb]{0.6,0.0,0.1}{jsrim: #1}}}
\newcommand{\change}[1]{{\color{red}#1}}
\newcommand{\changed}[1]{{\color{red}#1}}
\newcommand{\bb}[1]{\textbf{\textit{#1}}}

\renewcommand{\topfraction}{0.95}
\setcounter{bottomnumber}{1}
\renewcommand{\bottomfraction}{0.95}
\setcounter{totalnumber}{3}
\renewcommand{\textfraction}{0.05}
\renewcommand{\floatpagefraction}{0.95}
\setcounter{dbltopnumber}{2}
\renewcommand{\dbltopfraction}{0.95}
\renewcommand{\dblfloatpagefraction}{0.95}

\newcommand{\Net}[1]{#1}
\newcommand{\Loss}[1]{$\mathcal{L}_{#1}$}
\newcommand{\cm}{\checkmark}
\newcommand{\ts}{\textsuperscript}
\newcommand\oast{\stackMath\mathbin{\stackinset{c}{0ex}{c}{0ex}{\ast}{\bigcirc}}}

\makeatletter
\newcommand{\StatexIndent}[1][3]{%
  \setlength\@tempdima{\algorithmicindent}%
  \Statex\hskip\dimexpr#1\@tempdima\relax}
\makeatother

\newdimen{\algindent}
\setlength\algindent{1.5em}
\algnewcommand\LeftComment[2]{%
\hspace{#1\algindent}$\triangleright$ \eqparbox{COMMENT}{#2} \hfill %
}
\newcommand*\pct{\protect\scalebox{0.9}{\%}}
\newcommand*\smalleq{\protect\scalebox{0.9}{=}}
\newcommand*\MAES{\protect\scalebox{0.6}{${(\!\times\!10^{\texttt{-}1})}$}}
\def\nespace{\hskip\fontdimen2\font\relax}
\newcommand{\midsepremove}{\aboverulesep = 0mm \belowrulesep = 0mm}
\midsepremove
\newcommand{\midsepdefault}{\aboverulesep = 0.605mm \belowrulesep = 0.984mm}
\midsepdefault

\title{Iterative Filter Adaptive Network for 
Single Image Defocus Deblurring} 

\author{
    \vspace*{-10pt}\\
    Junyong Lee
    \qquad
    Hyeongseok Son
    \qquad
    Jaesung Rim
    \qquad
    Sunghyun Cho
    \qquad
    Seungyong Lee\\
    \vspace*{-11pt}\\
    POSTECH\\
    {\tt\small
    \{junyonglee, sonhs, jsrim123, s.cho, leesy\}@postech.ac.kr}\\
    {\tt\small \url{https://github.com/codeslake/IFAN}}
    \vspace*{5pt}
}

\maketitle
\begin{abstract}
We propose a novel end-to-end learning-based approach for single image defocus deblurring. The proposed approach is equipped with a novel Iterative Filter Adaptive Network (IFAN) that is specifically designed to handle spatially-varying and large defocus blur. For adaptively handling spatially-varying blur, IFAN predicts pixel-wise deblurring filters, which are applied to defocused features of an input image to generate deblurred features. For effectively managing large blur, IFAN models deblurring filters as stacks of small-sized separable filters. Predicted separable deblurring filters are applied to defocused features using a novel Iterative Adaptive Convolution (IAC) layer. We also propose a training scheme based on defocus disparity estimation and reblurring, which significantly boosts the deblurring quality. We demonstrate that our method achieves state-of-the-art performance both quantitatively and qualitatively on real-world images.

 \end{abstract} 
\section{Introduction} \label{sec:Intorduction}

Defocus deblurring aims to restore an all-in-focus image from a defocused image, and is highly demanded by daily photographers to remove unwanted blur.
Moreover, restored all-in-focus images can greatly facilitate high-level vision tasks such as semantic segmentation \cite{Noh:2015:ICCV, Wang:2020:DSRSS} and object detection \cite{Dai:2016:RFCN, Cao:2020:HQOD}.
Despite the usefulness, defocus deblurring remains a challenging problem as defocus blur is spatially varying in size, and its shape also varies across the image.

A conventional strategy \cite{Shi:2015:JNB, Andres:2016:RTF, Park:2017:unified, Cho:2017:Convergence, Karaali:2018:DMEAdaptive, Lee:2019:DMENet} is to model defocus blur as a combination of different convolution results obtained by applying predefined kernels to a sharp image.
It estimates per-pixel blur kernels based on the blur model and then performs non-blind deconvolution \cite{Fish:95:BD, Levin:2007:Coded, Krishnan:2008:deconvolution}.
However, this approach often fails due to the restrictive blur model, which disregards the nonlinearity of a real-world blur and constrains defocus blur in specific shapes such as disc \cite{Andres:2016:RTF} or Gaussian kernels \cite{Shi:2015:JNB, Park:2017:unified, Karaali:2018:DMEAdaptive, Lee:2019:DMENet}.
    
Recently, Abuolaim and Brown \cite{Abuolaim:2020:DPDNet} proposed the first end-to-end learning-based method, DPDNet, which does not rely on a specific blur model, but directly restores a sharp image.
Thanks to the end-to-end learning, DPDNet outperforms previous approaches for real-world defocused images.
They show that dual-pixel data obtainable from some modern cameras can significantly boost the deblurring performance, and present a dual-pixel defocus deblurring (DPDD) dataset.
However, ringing artifacts and remaining blur can often be found in their results, mainly due to its na\"ive UNet \cite{Ronneberger:2015:UNet} architecture, which is not flexible enough to deal with spatially-varying and large blur \cite{Zhou:2019:STFAN}.

In this paper, we propose an end-to-end network embedded with our novel \textit{Iterative Filter Adaptive Network (IFAN)} for single image defocus deblurring.
IFAN is specifically designed for the effective handling of spatially-varying and large defocus blur.
To handle the spatially-varying nature of defocus blur,
IFAN adopts an adaptive filter prediction scheme motivated by recent filter adaptive networks (FANs) \cite{zhang:2018:SVRNN,Zhou:2019:STFAN}. %
Specifically, IFAN does not directly predict pixel values, but generates spatially-adaptive per-pixel deblurring filters, which are then applied to features from an input defocus blurred image to generate deblurred features.

To efficiently handle large defocus blur that requires large receptive fields,
IFAN predicts stacks of small-sized separable filters instead of conventional filters unlike previous FANs.
To apply predicted separable filters to features, we also propose a novel \emph{Iterative Adaptive Convolution (IAC)} layer that iteratively applies separable filters to features.
As a result, IFAN significantly improves the deblurring quality at a low computational cost in the presence of spatially-varying and large defocus blur.

To further improve the single image deblurring quality, we train our network with novel defocus-specific tasks: defocus disparity estimation and reblurring.
The learning of defocus disparity estimation exploits dual-pixel data,
which provides stereo images with a tiny baseline, whose disparities are proportional to defocus blur magnitudes \cite{Garg:2019:DP, punnappurath:2020:modeling, Abuolaim:2020:DPDNet}.
Leveraging dual-pixel stereo images, we train IFAN to predict the disparity map from a single image so that it can also learn to predict blur magnitudes more accurately.

On the other hand, the learning of reblurring task, which is motivated by the reblur-to-deblur scheme in \cite{Chen:2018:R2D}, utilizes deblurring filters predicted by IFAN for reblurring all-in-focus images.
For accurate reblurring, IFAN needs to predict deblurring filters that contain accurate information about the shapes and sizes of defocus blur.
During training, we introduce an additional network that inverts predicted deblurring filters to reblurring filters and reblurs the ground-truth all-in-focus image.
We then train IFAN to minimize the difference between the defocused input image and the corresponding reblurred image.
We experimentally show that both tasks significantly boost the deblurring quality.

To verify the effectiveness of our method on diverse real-world images from different cameras, we extensively evaluate the method on several real-world datasets such as the DPDD dataset \cite{Abuolaim:2020:DPDNet}, Pixel dual-pixel test set \cite{Abuolaim:2020:DPDNet}, and CUHK blur detection dataset \cite{Shi:2014:CUHK}.
In addition, for quantitative evaluation, we present the Real Depth of Field (RealDOF) test set that provides real-world defocused images and their ground-truth all-in-focus images.

To summarize, our contributions include:
\vspace{-5pt}
\begin{itemize}
    \setlength\itemsep{0.01em}
    \item \textit{Iterative Filter Adaptive Network (IFAN)} that effectively handles spatially-varying and large defocus blur,
    \item a novel training scheme that utilizes the learning of defocus disparity estimation and reblurring, and
    \item state-of-the-art performance of defocus deblurring in terms of deblurring accuracy and computational cost. %
\end{itemize}

\section{Related Work}
\paragraph{Defocus deblurring}
Most defocus deblurring approaches,
including classical and recent deep-learning-based ones,
assume a specific blur model and try to estimate a defocus map based on the model.
They estimate the defocus map of an image leveraging various cues, such as
hand-crafted features \cite{Shi:2015:JNB, Krishnan:2008:deconvolution, Cho:2017:Convergence, Karaali:2018:DMEAdaptive}, learned regression features \cite{Andres:2016:RTF}, and deep features \cite{Park:2017:unified,Lee:2019:DMENet}.
They then utilize non-blind deconvolution methods such as \cite{Fish:95:BD, Levin:2007:Coded, Krishnan:2008:deconvolution} to restore a sharp image.
However, their performance is limited due to their restrictive blur models.
Recently, Abuolaim and Brown \cite{Abuolaim:2020:DPDNet} proposed an end-to-end learning-based approach that outperforms previous blur model-based ones.
However, their results often have ringing artifacts and remaining blur as mentioned in \Sec{\ref{sec:Intorduction}}.

\vspace{-13pt}
\paragraph{Filter adaptive networks}
FANs have been proposed to facilitate the spatially-adaptive handling of features in various tasks \cite{Jia:2016:DFN, Niklaus:2017:VFI, Niklaus:2017:VFISPC, Mildenhall:2018:BDKPN, Jo:2018:DUFSR, Wang:2018:RRTSR, Zhou:2019:STFAN, zhang:2018:SVRNN, Su:2019:PAC, Xu:2020:squeezesegv3}.
FANs commonly consist of two components: prediction of spatially-adaptive filters and transformation of features using the predicted filters, where the latter component is called filter adaptive convolution (FAC).
Various FANs have been proposed and applied to different tasks, such as frame interpolation \cite{Jia:2016:DFN, Niklaus:2017:VFI, Niklaus:2017:VFISPC}, denoising \cite{Mildenhall:2018:BDKPN}, super-resolution \cite{Jo:2018:DUFSR, Wang:2018:RRTSR}, semantic segmentation \cite{Su:2019:PAC}, and point cloud segmentation \cite{Xu:2020:squeezesegv3}.
For the deblurring task, Zhang \etal \cite{zhang:2018:SVRNN} proposed pixel-recurrent adaptive convolution for motion deblurring.
However, their method requires massive computational cost as the recurrent neural network must run for each pixel.
Zhou \etal \cite{Zhou:2019:STFAN} proposed a novel filter adaptive convolution layer for frame alignment and video deblurring.
However, handling large motion blur requires predicting large filters, which results in huge computational cost.

\begin{figure}[t]
\begin{center}
\includegraphics [width=0.8\linewidth] {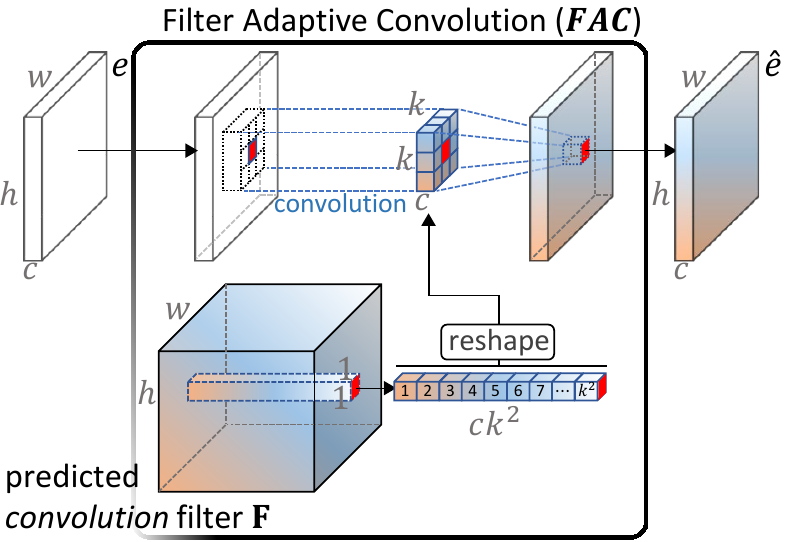}
\end{center}

\vspace{-0.4cm}
\caption{Filter Adaptive Convolution (FAC) \cite{Zhou:2019:STFAN}}
\label{fig:FAC}
\vspace{-14pt}
\end{figure}
\section{Single Image Defocus Deblurring}
In this section, we first introduce the Iterative Adaptive Convolution (IAC) layer, which forms the basis of our IFAN (\Sec{\ref{sec:IAC}}).
Then, we present our deblurring network based on IFAN with detailed explanations on each component (\Sec{\ref{sec:Deblurring_Newtwork}}).
Finally, we explain our training strategy exploiting the disparity estimation and reblurring tasks (\Sec{\ref{sec:training}}).

\subsection{Iterative Adaptive Convolution (IAC)}\label{sec:IAC}

Our IAC is inspired by the filter adaptive convolution (FAC) \cite{Zhou:2019:STFAN} that forms the basis of previous FANs.
For a better understanding of IAC, we first briefly review FAC.

\Fig{\ref{fig:FAC}} shows an overview of a FAC layer.
A FAC layer takes a feature map $e$ and a map of spatially-varying convolution filters $\textbf{F}$ as input.
$e$ and $\textbf{F}$ have the same spatial size $h\!\times\!w$.
The input feature map $e$ has $c$ channels.
At each spatial location in $\textbf{F}$ is a $ck^2$-dim vector representing $c$ convolution filters of size $k\!\times\!k$.
For each spatial location $(x, y)$, the FAC layer generates $c$ convolution filters from $\textbf{F}$ by reshaping the vector at $(x, y)$.
Then, the layer applies the filters to the features in $e$ centered at $(x,y)$ in the channel-wise manner to generate an output feature map $\hat{e} \in \mathbb{R}^{h\times w \times c}$.

For effectively handling spatially-varying and large defocus blur, it is critical to secure large receptive fields.
However, while FAC facilitates spatially-adaptive processing of features, increasing the filter size to cover wider receptive fields results in huge memory consumption and computational cost.
To resolve this limitation, we propose the IAC layer that iteratively applies small-sized separable filters to efficiently enlarge the receptive fields with little computational overhead.

\begin{figure}[t]
\begin{center}
\includegraphics [width=0.8\linewidth] {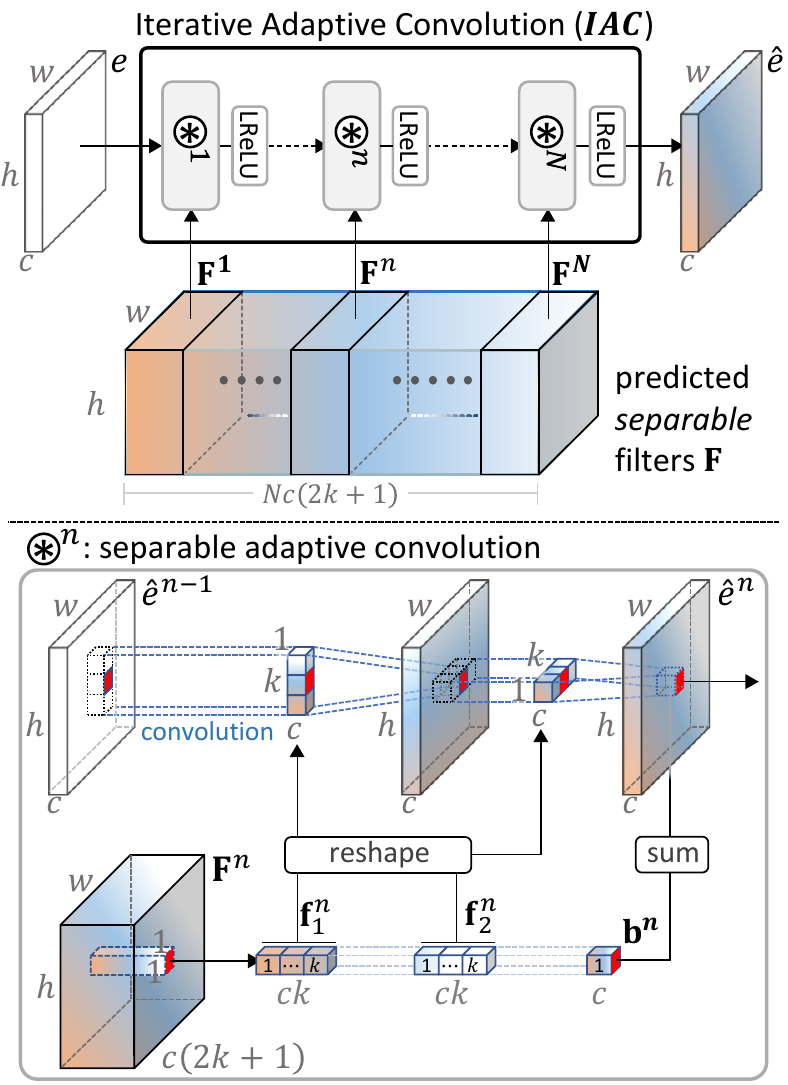}
\end{center}
\vspace{-0.4cm}
\caption{
Proposed Iterative Adaptive Convolution (IAC)
}
\label{fig:IAC}
\vspace{-11pt}
\end{figure}

\Fig{\ref{fig:IAC}} shows an overview of our IAC layer.
Similarly to FAC, IAC takes a feature map $e$ and a map of spatially-varying filters $\textbf{F}$ as input, whose spatial sizes are the same.
At each spatial location in $\textbf{F}$ is a $Nc(2k\,\texttt{+}\,1)$-dim vector representing $N$ sets of filters $\{\textbf{F}^1,\textbf{F}^2,\cdots,\textbf{F}^N\}$.
The $n$-th filter set $\textbf{F}^n$ has two 1-dim filters $\textbf{f}_1^{\,n}$ and $\textbf{f}_2^{\,n}$ of the sizes $k\!\times\!1$ and $1\!\times\!k$, respectively, and one bias vector $\textbf{b}^n\!$.
$\ \textbf{f}_1^{\,n}$, $\textbf{f}_2^{\,n}$, and $\textbf{b}^n$ have $c$ channels.
The IAC layer decomposes
the vector
in $\textbf{F}$ at each location into filters and bias vectors, and iteratively applies them to $e$ in the channel-wise manner to produce an output feature map $\hat{e}$.

Let $\hat{e}^n$ denote the $n$-th intermediate feature map after applying the $n$-th separable filters and bias, where $\hat{e}^0\,\texttt{=}\,\,e$ and $\hat{e}^N\texttt{=}\,\,\hat{e}$.
Then, the IAC layer computes $\hat{e}^n$ for $n\in\{1,\cdots,N\}$ as follows:
\begin{equation}
    \hat{e}^n = \textrm{LReLU}(\hat{e}^{(n-1)}\,*\, \textbf{f}_1^{\,n}\,*\,\textbf{f}_2^{\,n} + \textbf{b}^n),
\end{equation}
where $\textrm{LReLU}$ is the leaky rectified linear unit \cite{Maas:2013:Leaky} and $*$ is the channel-wise convolution operator that performs convolutions in a spatially-adaptive manner.

Separable filters in our IAC layer play a key role in resolving the limitation of the FAC layer. 
Xu \etal \cite{Xu:2014:Deep} showed that a convolutional network with 1-dim filters can successfully approximate a large inverse filter for the deconvolution task.
Similarly, our IAC layer secures larger receptive fields at much lower memory and computational costs than the FAC layer by utilizing 1-dim filters, instead of 2-dim convolutions.
However, compared to dense 2-dim convolution filters in the FAC layer, our separable filters may not provide enough accuracy for deblurring filters.
We handle this problem by iteratively applying separable filters to fully exploit the nonlinear nature of a deep network. 
Our iterative scheme also enables small-sized separable filters to be used for establishing large receptive fields.

\begin{figure*}[t]
\begin{center}
\includegraphics [width=1.0\linewidth] {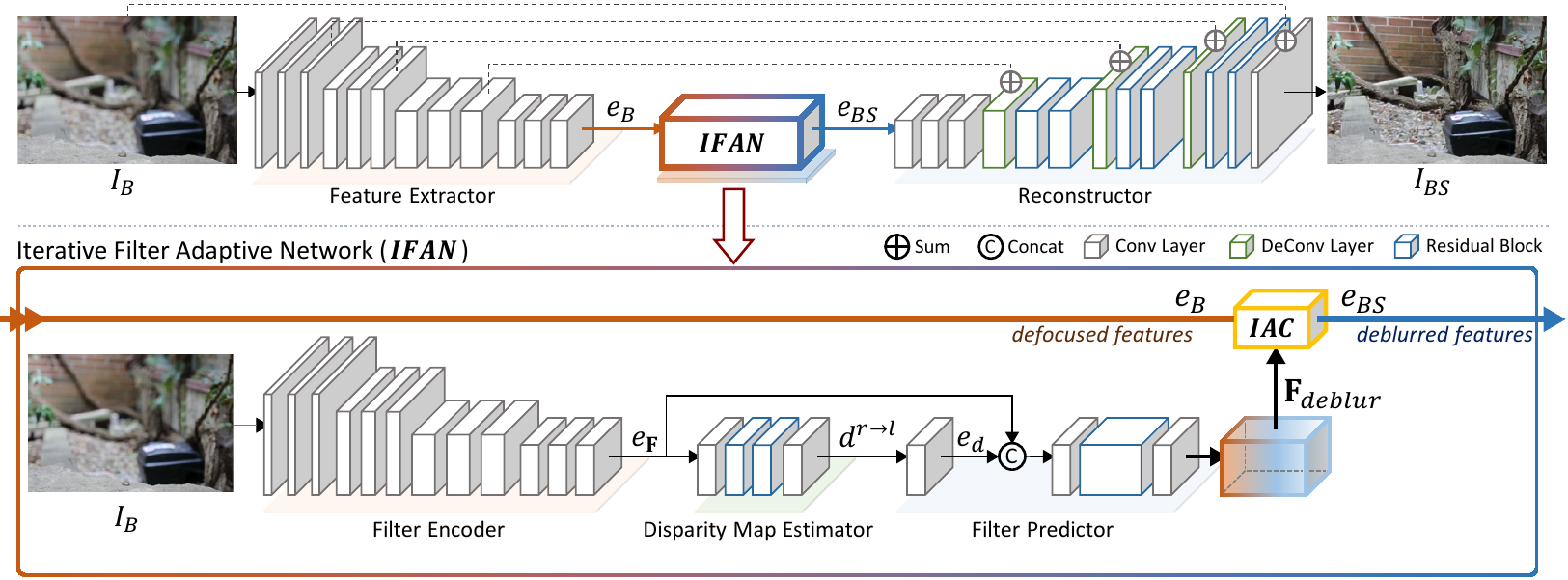}
\end{center}
\vspace{-0.4cm}
  \caption{Proposed defocus deblurring network with Iterative Filter Adaptive Network (IFAN).
  }
\label{fig:network}
\vspace{-12pt}
\end{figure*}
\subsection{Deblurring Network with IFAN}
\label{sec:Deblurring_Newtwork}

\Fig{\ref{fig:network}} shows an overview of our deblurring network based on IFAN.
Our network takes a single defocused image $I_B\!\in\!\mathbb{R}^{H\times W\times3 }$ as input and produces a deblurred output $I_{BS}\!\in\!\mathbb{R}^{H\times W\times 3}$,
where $H$ and $W$ are the height and width of an image, respectively.
The network is built upon a simple encoder-decoder architecture consisting of a feature extractor, reconstructor, and IFAN module in the middle.
The feature extractor extracts defocused features $e_B\!\in\! \mathbb{R}^{h\times w\times c_{e}}$,
where $h~\texttt{=}\,\frac{H}{8}$ and $w~\texttt{=}\,\frac{W}{8}$,
and feeds them to IFAN.
IFAN removes defocus blur in the feature domain by predicting spatially-varying deblurring filters and applying them to $e_B$ using IAC.
The deblurred features $e_{BS}$ from IFAN is then passed to the reconstructor, which restores an all-in-focus image $I_{BS}$.
The detailed architecture of each network component
can be found in our supplementary material.
In the following, we describe IFAN in more detail.

\vspace{-13pt}
\paragraph{Iterative Filter Adaptive Network (IFAN)}\label{ssec:IFAN}
IFAN takes defocused features $e_B$ as input, and produces deblurred features $e_{BS}$.
To produce deblurred features, IFAN takes $I_B$ as additional input, and predicts deblurring filters from $I_B$. 
IFAN consists of a filter encoder, disparity map estimator, filter predictor, and IAC layer.
The filter encoder encodes $I_B$ into $e_\textbf{F}\!\in\! \mathbb{R}^{h\times w\times c_{e}}$, which is then passed to the disparity map estimator.
The disparity map estimator is a sub-network specifically designed to exploit dual-pixel data for effectively training IFAN, which will be discussed in \Sec{\ref{sec:training}}.
After the disparity map estimator, the filter predictor predicts a deblurring filter map $\textbf{F}_{deblur} \!\in\! \mathbb{R}^{h\times w\times c_{\textbf{F}_{deblur}}}$, where $c_{\textbf{F}_{deblur}}\texttt{=}\,Nc_{e}(2k\,\texttt{+}\,1)$.
Finally, the IAC layer transforms the input features $e_B$ using the predicted filters $\textbf{F}_{deblur}$ to generate deblurred features $e_{BS}$.

\subsection{Network Training}
\label{sec:training}

We train our network using the DPDD dataset with two defocus-specific tasks: defocus disparity estimation and reblurring.
In this section, we explain the training data, our training strategy using each task, and our final loss function.

\vspace{-13pt}
\paragraph{Dataset}
We use dual-pixel images from the DPDD dataset \cite{Abuolaim:2020:DPDNet} to train our network.
A dual-pixel image provides a pair of stereo images with a tiny baseline, whose disparities are proportional to defocus blur magnitudes.
The DPDD dataset provides 500 dual-pixel images captured with a Canon EOS 5D Mark IV.
Each dual-pixel image is provided in the form of a pair of left and right stereo images $I_B^l$ and $I_B^r$, respectively. %
For each dual-pixel image, the dataset also provides a defocused image $I_B$, which is generated by merging $I_B^l$ and $I_B^r$, and its corresponding ground-truth all-in-focus image $I_S$.
The 500 dual-pixel images are split into training, validation, and testing sets, each of which contains 350, 74, and 76 scenes, respectively.
Refer to \cite{Abuolaim:2020:DPDNet} for more details on the DPDD dataset.
\begin{figure}[t]
\begin{center}
\includegraphics [width=0.9\linewidth] {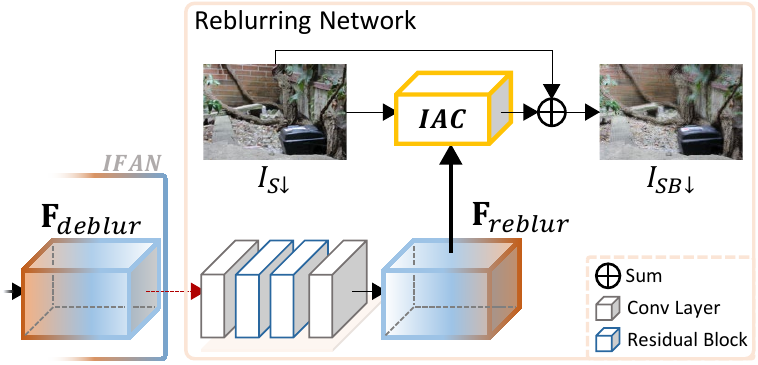}
\end{center}
\vspace{-0.4cm}
\caption{The reblurring network}
\label{fig:RBN}
\vspace{-14pt}
\end{figure}
\begin{table*}[t]
\centering

\small
\begin{tabularx}{\textwidth}{c@{\hspace{25pt}}c Y c Y c@{\hspace{30pt}}c@{\hspace{24pt}}c@{\hspace{24pt}}c Y c@{\hspace{20pt}}c}
\toprule

\multicolumn{2}{c}{\Net{IFAN}} && \multirow{2}{*}[-0.5\dimexpr \aboverulesep + \belowrulesep + \cmidrulewidth]{\makecell{RBN}}&& \multicolumn{4}{c}{Evaluations on the DPDD Dataset \cite{Abuolaim:2020:DPDNet}} && \multicolumn{2}{c}{Computational Costs}
\\
\cmidrule(lr){1-2} \cmidrule(lr){6-9} \cmidrule(lr){11-12}
$\text{FP} + \text{IAC}$ & DME &&  %
&& PSNR$\uparrow$\! & SSIM$\uparrow$\! & MAE\MAES$\downarrow$\! & LPIPS$\downarrow$\! && Params (M) & MACs (B)
\\
\midrule
\arrayrulecolor{lightgray}
&     &&         && 24.88 & 0.753 & 0.416 & 0.289 && \multirow{2}{*}{10.58} & \multirow{2}{*}{364.3} %
\\
& \cm &&         && 24.97 & 0.761 & 0.412 & 0.280 && &  \\ %
\cmidrule{1-12}
\cm &     &&     && 25.07 & 0.765 & 0.406 & 0.271 && \multirow{4}{*}{10.48} & \multirow{4}{*}{362.9}
\\
\cm & \cm &&     && 25.18 & 0.780 & 0.403 & 0.233 && 
\\ 
\cm &     && \cm && 25.28 & 0.780 & 0.400 &  0.245 &&
\\ 
\cm & \cm && \cm && \textbf{25.37} & \textbf{0.789} & \textbf{0.394} & \textbf{0.217} && %
\\
\arrayrulecolor{black}
\bottomrule
\end{tabularx}

\vspace{0.1cm}
\caption{
An ablation study.
FP, DME, and RBN indicate the filter predictor, disparity map estimator, and reblurring network, respectively.
The first row corresponds to the baseline model. For fair evaluation, to obtain the baseline model, the components of our model are replaced by conventional convolution layers and residual blocks with similar model parameter numbers and computational costs.
}
\label{tbl:ablation}
\vspace*{-5pt}
\end{table*}

\begin{figure*}[t]
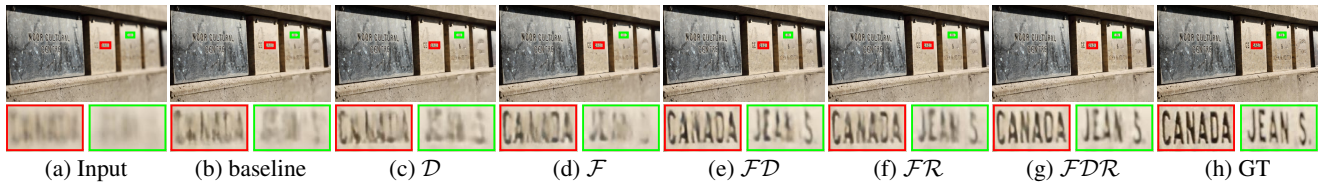

    \centering
    \def \ext {.jpg} %

    \def \corl {0.579} %
    \def \corb {0.544} %
    \def \corrt {0.08} %
    
    \def \corls {0.738} %
    \def \corbs {0.655} %
    \def \corrts {0.07} %

    \def \img {70\ext}
    
    \def \wb {0.122}
    \def \ws {0.059}

    \setlength\tabcolsep{0.8 pt}
    \small
    \begin{tabular}{l@{\hspace{0pt}}rl@{\hspace{0pt}}rl@{\hspace{0pt}}rl@{\hspace{0pt}}rl@{\hspace{0pt}}rl@{\hspace{0pt}}rl@{\hspace{0pt}}rl@{\hspace{0pt}}r}
    
    \multicolumn{2}{c}{
    \begin{tikzpicture}
    \node[anchor=south west,inner sep=0] (image) at (0,0) {\includegraphics[trim={0 50px 0 50px},clip, width=\wb\textwidth]{data/experiments/ablation/figures/Input/\img}};
    \begin{scope}[x={(image.south east)},y={(image.north west)}]
        \draw[red,thick] ([shift={(0.01, 0.01)}]\corl, \corb) rectangle ([shift={(-0.01, -0.01)}]{\corl+\corrt}, {\corb+\corrt});
        \draw[green,thick] ([shift={(0.01, 0.01)}]\corls, \corbs) rectangle ([shift={(-0.01, -0.01)}]{\corls+\corrts}, {\corbs+\corrts});
    \end{scope}
    \end{tikzpicture}} &
    
    \multicolumn{2}{c}{
    \begin{tikzpicture}
    \node[anchor=south west,inner sep=0] (image) at (0,0) {\includegraphics[trim={0 50px 0 50px},clip, width=\wb\textwidth]{data/experiments/ablation/figures/B/\img}};
    \begin{scope}[x={(image.south east)},y={(image.north west)}]
        \draw[red,thick] ([shift={(0.01, 0.01)}]\corl, \corb) rectangle ([shift={(-0.01, -0.01)}]{\corl+\corrt}, {\corb+\corrt});
        \draw[green,thick] ([shift={(0.01, 0.01)}]\corls, \corbs) rectangle ([shift={(-0.01, -0.01)}]{\corls+\corrts}, {\corbs+\corrts});
    \end{scope}
    \end{tikzpicture}} &

    \multicolumn{2}{c}{
    \begin{tikzpicture}
    \node[anchor=south west,inner sep=0] (image) at (0,0) {\includegraphics[trim={0 50px 0 50px},clip,width=\wb\textwidth]{data/experiments/ablation/figures/DPD/D/\img}};
    \begin{scope}[x={(image.south east)},y={(image.north west)}]
        \draw[red,thick] ([shift={(0.01, 0.01)}]\corl, \corb) rectangle ([shift={(-0.01, -0.01)}]{\corl+\corrt}, {\corb+\corrt});
        \draw[green,thick] ([shift={(0.01, 0.01)}]\corls, \corbs) rectangle ([shift={(-0.01, -0.01)}]{\corls+\corrts}, {\corbs+\corrts});
    \end{scope}
    \end{tikzpicture}} &
    
    \multicolumn{2}{c}{
    \begin{tikzpicture}
    \node[anchor=south west,inner sep=0] (image) at (0,0) {\includegraphics[trim={0 50px 0 50px},clip,width=\wb\textwidth]{data/experiments/ablation/figures/DPD/S/\img}};
    \begin{scope}[x={(image.south east)},y={(image.north west)}]
        \draw[red,thick] ([shift={(0.01, 0.01)}]\corl, \corb) rectangle ([shift={(-0.01, -0.01)}]{\corl+\corrt}, {\corb+\corrt});
        \draw[green,thick] ([shift={(0.01, 0.01)}]\corls, \corbs) rectangle ([shift={(-0.01, -0.01)}]{\corls+\corrts}, {\corbs+\corrts});
    \end{scope}
    \end{tikzpicture}} &
    
    \multicolumn{2}{c}{
    \begin{tikzpicture}
    \node[anchor=south west,inner sep=0] (image) at (0,0) {\includegraphics[trim={0 50px 0 50px},clip,width=\wb\textwidth]{data/experiments/ablation/figures/DPD/SD/\img}};
    \begin{scope}[x={(image.south east)},y={(image.north west)}]
        \draw[red,thick] ([shift={(0.01, 0.01)}]\corl, \corb) rectangle ([shift={(-0.01, -0.01)}]{\corl+\corrt}, {\corb+\corrt});
        \draw[green,thick] ([shift={(0.01, 0.01)}]\corls, \corbs) rectangle ([shift={(-0.01, -0.01)}]{\corls+\corrts}, {\corbs+\corrts});
    \end{scope}
    \end{tikzpicture}} &
    
    \multicolumn{2}{c}{
    \begin{tikzpicture}
    \node[anchor=south west,inner sep=0] (image) at (0,0) {\includegraphics[trim={0 50px 0 50px},clip,width=\wb\textwidth]{data/experiments/ablation/figures/DPD/SR/\img}};
    \begin{scope}[x={(image.south east)},y={(image.north west)}]
        \draw[red,thick] ([shift={(0.01, 0.01)}]\corl, \corb) rectangle ([shift={(-0.01, -0.01)}]{\corl+\corrt}, {\corb+\corrt});
        \draw[green,thick] ([shift={(0.01, 0.01)}]\corls, \corbs) rectangle ([shift={(-0.01, -0.01)}]{\corls+\corrts}, {\corbs+\corrts});
    \end{scope}
    \end{tikzpicture}} &
    
    \multicolumn{2}{c}{
    \begin{tikzpicture}
    \node[anchor=south west,inner sep=0] (image) at (0,0) {\includegraphics[trim={0 50px 0 50px},clip,width=\wb\textwidth]{data/experiments/ablation/figures/DPD/SDR/\img}};
    \begin{scope}[x={(image.south east)},y={(image.north west)}]
        \draw[red,thick] ([shift={(0.01, 0.01)}]\corl, \corb) rectangle ([shift={(-0.01, -0.01)}]{\corl+\corrt}, {\corb+\corrt});
        \draw[green,thick] ([shift={(0.01, 0.01)}]\corls, \corbs) rectangle ([shift={(-0.01, -0.01)}]{\corls+\corrts}, {\corbs+\corrts});
    \end{scope}
    \end{tikzpicture}} &

    \multicolumn{2}{c}{
    \begin{tikzpicture}
    \node[anchor=south west,inner sep=0] (image) at (0,0) {\includegraphics[trim={0 50px 0 50px},clip,width=\wb\textwidth]{data/experiments/ablation/figures/GT/\img}};
    \begin{scope}[x={(image.south east)},y={(image.north west)}]
        \draw[red,thick] ([shift={(0.01, 0.01)}]\corl, \corb) rectangle ([shift={(-0.01, -0.01)}]{\corl+\corrt}, {\corb+\corrt});
        \draw[green,thick] ([shift={(0.01, 0.01)}]\corls, \corbs) rectangle ([shift={(-0.01, -0.01)}]{\corls+\corrts}, {\corbs+\corrts});
    \end{scope}
    \end{tikzpicture}} \\[-0.03in]
    
    \adjustbox{width=\ws\textwidth, trim={\corl\width} {\corb\height} {\width-(\corl\width+\corrt\width)} {\height-(\corb\height+\corrt\height)},clip,cfbox=red 0.7pt -0.7pt}
    {\includegraphics[trim={0 50px 0 50px},clip,width=\ws\textwidth]{data/experiments/ablation/figures/Input/\img}} &
    \adjustbox{width=\ws\textwidth, trim={\corls\width} {\corbs\height} {\width-(\corls\width+\corrts\width)} {\height-(\corbs\height+\corrts\height)}, clip,cfbox=green 0.7pt -0.7pt}
    {\includegraphics[trim={0 50px 0 50px},clip,width=\ws\textwidth]{data/experiments/ablation/figures/Input/\img}} &
    
    \adjustbox{width=\ws\textwidth, trim={\corl\width} {\corb\height} {\width-(\corl\width+\corrt\width)} {\height-(\corb\height+\corrt\height)},clip,cfbox=red 0.7pt -0.7pt}
    {\includegraphics[trim={0 50px 0 50px},clip,width=\ws\textwidth]{data/experiments/ablation/figures/B/\img}} &
    \adjustbox{width=\ws\textwidth, trim={\corls\width} {\corbs\height} {\width-(\corls\width+\corrts\width)} {\height-(\corbs\height+\corrts\height)}, clip,cfbox=green 0.7pt -0.7pt}
    {\includegraphics[trim={0 50px 0 50px},clip,width=\ws\textwidth]{data/experiments/ablation/figures/B/\img}} &
    
    \adjustbox{width=\ws\textwidth, trim={\corl\width} {\corb\height} {\width-(\corl\width+\corrt\width)} {\height-(\corb\height+\corrt\height)}, clip,cfbox=red 0.7pt -0.7pt}
    {\includegraphics[trim={0 50px 0 50px},clip,width=\ws\textwidth]{data/experiments/ablation/figures/DPD/D/\img}} &
    \adjustbox{width=\ws\textwidth, trim={\corls\width} {\corbs\height} {\width-(\corls\width+\corrts\width)} {\height-(\corbs\height+\corrts\height)}, clip,cfbox=green 0.7pt -0.7pt}
    {\includegraphics[trim={0 50px 0 50px},clip,width=\ws\textwidth]{data/experiments/ablation/figures/DPD/D/\img}} &

    \adjustbox{width=\ws\textwidth, trim={\corl\width} {\corb\height} {\width-(\corl\width+\corrt\width)} {\height-(\corb\height+\corrt\height)}, clip,cfbox=red 0.7pt -0.7pt}
    {\includegraphics[trim={0 50px 0 50px},clip,width=\ws\textwidth]{data/experiments/ablation/figures/DPD/S/\img}} &
    \adjustbox{width=\ws\textwidth, trim={\corls\width} {\corbs\height} {\width-(\corls\width+\corrts\width)} {\height-(\corbs\height+\corrts\height)}, clip,cfbox=green 0.7pt -0.7pt}
    {\includegraphics[trim={0 50px 0 50px},clip,width=\ws\textwidth]{data/experiments/ablation/figures/DPD/S/\img}} &
    
    \adjustbox{width=\ws\textwidth, trim={\corl\width} {\corb\height} {\width-(\corl\width+\corrt\width)} {\height-(\corb\height+\corrt\height)}, clip,cfbox=red 0.7pt -0.7pt}
    {\includegraphics[trim={0 50px 0 50px},clip,width=\ws\textwidth]{data/experiments/ablation/figures/DPD/SD/\img}} &
    \adjustbox{width=\ws\textwidth, trim={\corls\width} {\corbs\height} {\width-(\corls\width+\corrts\width)} {\height-(\corbs\height+\corrts\height)}, clip,cfbox=green 0.7pt -0.7pt}
    {\includegraphics[trim={0 50px 0 50px},clip,width=\ws\textwidth]{data/experiments/ablation/figures/DPD/SD/\img}} &
    
    \adjustbox{width=\ws\textwidth, trim={\corl\width} {\corb\height} {\width-(\corl\width+\corrt\width)} {\height-(\corb\height+\corrt\height)}, clip,cfbox=red 0.7pt -0.7pt}
    {\includegraphics[trim={0 50px 0 50px},clip,width=\ws\textwidth]{data/experiments/ablation/figures/DPD/SR/\img}} &
    \adjustbox{width=\ws\textwidth, trim={\corls\width} {\corbs\height} {\width-(\corls\width+\corrts\width)} {\height-(\corbs\height+\corrts\height)}, clip,cfbox=green 0.7pt -0.7pt}
    {\includegraphics[trim={0 50px 0 50px},clip,width=\ws\textwidth]{data/experiments/ablation/figures/DPD/SR/\img}} &
    
    \adjustbox{width=\ws\textwidth, trim={\corl\width} {\corb\height} {\width-(\corl\width+\corrt\width)} {\height-(\corb\height+\corrt\height)}, clip,cfbox=red 0.7pt -0.7pt}
    {\includegraphics[trim={0 50px 0 50px},clip,width=\ws\textwidth]{data/experiments/ablation/figures/DPD/SDR/\img}} &
    \adjustbox{width=\ws\textwidth, trim={\corls\width} {\corbs\height} {\width-(\corls\width+\corrts\width)} {\height-(\corbs\height+\corrts\height)}, clip,cfbox=green 0.7pt -0.7pt}
    {\includegraphics[trim={0 50px 0 50px},clip,width=\ws\textwidth]{data/experiments/ablation/figures/DPD/SDR/\img}} &

    \adjustbox{width=\ws\textwidth, trim={\corl\width} {\corb\height} {\width-(\corl\width+\corrt\width)} {\height-(\corb\height+\corrt\height)}, clip,cfbox=red 0.7pt -0.7pt}
    {\includegraphics[trim={0 50px 0 50px},clip,width=\ws\textwidth]{data/experiments/ablation/figures/GT/\img}} &
    \adjustbox{width=\ws\textwidth, trim={\corls\width} {\corbs\height} {\width-(\corls\width+\corrts\width)} {\height-(\corbs\height+\corrts\height)}, clip,cfbox=green 0.7pt -0.7pt}
    {\includegraphics[trim={0 50px 0 50px},clip,width=\ws\textwidth]{data/experiments/ablation/figures/GT/\img}}
    
    \\ [-0.02in]
    \multicolumn{2}{c}{(a) Input } &
    \multicolumn{2}{c}{(b) baseline} &
    \multicolumn{2}{c}{(c) $\mathcal{D}$} &
    \multicolumn{2}{c}{(d) $\mathcal{F}$} &
    \multicolumn{2}{c}{(e) $\mathcal{F}\mathcal{D}$} &
    \multicolumn{2}{c}{(f) $\mathcal{F}\mathcal{R}$} &
    \multicolumn{2}{c}{(g) $\mathcal{F}\mathcal{D}\mathcal{R}$} &
    \multicolumn{2}{c}{(h) GT}
  \end{tabular}%
  
  \vspace{0.05cm}
  \caption{Qualitative results of an ablation study on the DPDD dataset \cite{Abuolaim:2020:DPDNet}.
  The first and last columns show a defocused input image and its ground-truth all-in-focus image, respectively.
  Between the columns, the letters in each sub-caption indicate combination of components (refer \Tbl{\ref{tbl:ablation}}).
  $\mathcal{F}$ means the filter predictor and the IAC layer,
  $\mathcal{D}$ means the disparity map estimator, and
  $\mathcal{R}$ means the reblurring network.
  The baseline implies a model without $\mathcal{F}$, $\mathcal{D}$ and $\mathcal{R}$.
  Images in the red and green box are zoomed-in cropped patches. 
  }
  \vspace{-14pt}
  \label{fig:ablation}
\end{figure*}

\vspace{-13pt}
\paragraph{Defocus disparity estimation}
As the disparities between dual-pixel stereo images are proportional to blur magnitudes, we can train IFAN to learn more accurate defocus blur information by training IFAN to predict the disparity map.
To this end, during training, we feed a right image $I_B^r$ from a pair of dual-pixel stereo images to our deblurring network.
Then, we train the disparity map estimator to predict a disparity map $d^{r\rightarrow l}\!\in\!\mathbb{R}^{h\times w}$ between the downsampled left and right stereo images.
Specifically, we train the disparity map estimator using a disparity loss defined as:
\begin{equation}
    \mathcal{L}_{disp} = \emph{MSE}(I_{B\downarrow}^{r\rightarrow l}, I_{B\downarrow}^l),
    \label{eq:L_disp}
\end{equation}
where $\emph{MSE}(\cdot)$ is the mean-squared error function.
$I_{B\downarrow}^l$ is a left image downsampled by $\frac{1}{8}$.
$I_{B\downarrow}^{r\rightarrow l}$ is a right image downsampled by $\frac{1}{8}$ and warped by the disparity map $d^{r\rightarrow l}$.
We use the spatial transformer \cite{Jaderberg:2015:STN} for warping.
By minimizing $\mathcal{L}_{disp}$, both filter encoder and disparity map estimator are trained to predict an accurate disparity map and eventually accurate defocus magnitudes.
Note that we utilize dual-pixel images only for training, and our trained network requires only a single defocused image as its input.

\vspace{-13pt}
\paragraph{Reblurring}
We train IFAN also using the reblurring task.
For the learning of reblurring, we introduce an auxiliary reblurring network.
The reblurring network is attached at the end of IFAN and trained to invert deblurring filters $\textbf{F}_{deblur}$ to reblurring filters $\textbf{F}_{reblur}$ (\Fig{\ref{fig:RBN}}).
Then, using $\textbf{F}_{reblur}$, the IAC layer reblurs a downsampled ground-truth image $I_{S\downarrow}\!\in\!\mathbb{R}^{h\times w\times 3}$ to reproduce a downsampled version of the defocused input image.
For training IFAN as well as the reblurring network, we use a reblurring loss defined as:
\begin{equation}
    \mathcal{L}_{reblur}=\emph{MSE}(I_{SB\downarrow}, I_{B\downarrow}),
    \label{eq:L_reblur}
\end{equation}
where $I_{SB\downarrow}$ is a reblurred image obtained from $I_{S\downarrow}$ using $\textbf{F}_{reblur}$, and $I_{B\downarrow}$ is a downsampled input image.
$\mathcal{L}_{reblur}$ induces IFAN to predict $\textbf{F}_{deblur}$ containing valid information about blur shapes and sizes needed for accurate reblurring. Such information eventually improves the performance of deblurring filters used for the final defocus deblurring.
Note that we utilize the reblurring network only for training.

\vspace{-13pt}
\paragraph{Loss functions}
In addition to the disparity and reblurring losses, we use a deblurring loss, which is defined as:
\begin{equation}
\mathcal{L}_{deblur}=\emph{MSE}(I_{BS}, I_{S}).
\label{eq:deblur}
\end{equation}

Our total loss function to train our network is then defined as $\mathcal{L}_{total}\,\texttt{=}\,\mathcal{L}_{deblur}\,\texttt{+}~\mathcal{L}_{disp}\,\texttt{+}~\mathcal{L}_{reblur}$.
Each loss term affects different parts of our network.
While $\mathcal{L}_{deblur}$ trains the feature extractor, IFAN, and reconstructor,
$\mathcal{L}_{disp}$ trains only the filter encoder and disparity map estimator in IFAN.
$\mathcal{L}_{reblur}$ trains both IFAN and reblurring network.
Note that we use dual-pixel stereo images $(I_B^l, I_B^r)$ only for training.
Both $\mathcal{L}_{deblur}$ and $\mathcal{L}_{reblur}$ utilize $I_B$ while $\mathcal{L}_{disp}$ utilizes $(I_B^l, I_B^r)$.
In this way, we can fully utilize the DPDD dataset for training our network.

\begin{table*}[t]
\centering
\small
\begin{tabularx}{\textwidth}{
    X
    Y
    Y
    Y
    Y
    c@{\hspace{15pt}}
    c@{\hspace{15pt}}
    c}
\toprule

 \multirow{2}{*}[-0.5\dimexpr \aboverulesep + \belowrulesep + \cmidrulewidth]{Model} & \multicolumn{4}{c}{Evaluations the DPDD Dataset \cite{Abuolaim:2020:DPDNet}} & \multicolumn{3}{c}{Computational Costs}  \\
 
\cmidrule(lr){2-5} \cmidrule(lr){6-8}

& PSNR$\uparrow$\! & SSIM$\uparrow$\! & MAE\MAES$\downarrow$\! & LPIPS$\downarrow$\! & Params (M) & MACs (B) & Time (Sec) \\
\midrule
Input & 23.89 & 0.725 & 0.471 & 0.349 & - & - & - \\
\arrayrulecolor{lightgray}
\cmidrule{1-8}

JNB \cite{Shi:2015:JNB}
& 23.69 & 0.707 & 0.480 & 0.442 & - & - & 105.8\\ %

EBDB \cite{Karaali:2018:DMEAdaptive}
& 23.94 & 0.723 & 0.468 & 0.402 & - & - & 96.58\\ %

DMENet \cite{Lee:2019:DMENet}
& 23.90 & 0.720 & 0.470 & 0.410 & 26.94 & 1172.5 & 77.70\\ %

DPDNet$_{S}$ \cite{Abuolaim:2020:DPDNet}
& 24.03 & 0.735 & 0.461 & 0.279 & 35.25 & 989.8 & 0.462\\

DPDNet$_{D}$ \cite{Abuolaim:2020:DPDNet}
& 25.23 & 0.787 & 0.401 & 0.224 & 35.25 & 991.4 & 0.474\\ %

\cmidrule{1-8}
Ours
& \textbf{25.37} & \textbf{0.789} & \textbf{0.394} & \textbf{0.217} & \textbf{10.48} & \textbf{362.9} & \textbf{0.014}\\
\arrayrulecolor{black}
\bottomrule
\end{tabularx}

\vspace{0.1cm} 
\caption{Quantitative comparison with previous defocus deblurring methods.
All the methods are evaluated using the code provided by the authors.
JNB and EBDB are not deep learning-based methods, so their parameter numbers and MACs are not available.
DPDNet$_{D}$ \cite{Abuolaim:2020:DPDNet} takes not a single defocused image but dual-pixel stereo images as input at test time.
All the other methods, including ours, take a single defocused image as input at test time.
}
\label{tbl:comparison:DP}
\vspace{-5pt}
\end{table*}

\section{Experiments}

We implemented our models using PyTorch \cite{paszke2017automatic}.
We use rectified-Adam \cite{Liu:2020:RADAM} with $\beta_1\,\texttt{=}~0.9$, $\beta_2\,\texttt{=}~0.99$ and weight decay rate 0.01 for training our network.
We use gradient norm clipping with the value empirically set to 0.5.
The network is trained for 600k iterations with an initial learning rate of $1.0\!\times\!10^{\texttt{-}4}$, which is step-decayed to half at the 500k-th and 550k-th iterations, which are experimentally chosen.
We set the number of filters $N\,\texttt{=}\,17$ for $\textbf{F}_{deblur}$ and $\textbf{F}_{reblur}$.
For each iteration, we randomly sample a batch of images from the DPDD training set. 
We use a batch size of 8, and randomly augment images with Gaussian noise, gray-scale image conversion, and scaling, then crop them to $256\!\times\!256$.

For the evaluation of defocus deblurring performance, we measure the Peak Signal-to-Noise Ratio (PSNR), Structural Similarity (SSIM) \cite{wang2004ssim}, Mean Absolute Error (MAE), and Learned Perceptual Image Patch Similarity (LPIPS) \cite{zhang2018perceptual} between deblurred results and their corresponding ground-truth images.
We evaluate our models and previous ones on a PC with an NVIDIA GeForce RTX 3090 GPU.

\subsection{Ablation Study}
To analyze the effect of each component of our model, we conduct an ablation study (\Tbl{\ref{tbl:ablation}} and \Fig{\ref{fig:ablation}}).
All models in the ablation study are trained under the same conditions (\eg, optimizer, batch size, learning rate, random seeds, etc.).
For the validation, we compare a stripped-down baseline model and its five variants.
For the baseline model, we restructure the key modules specifically designed for our network using conventional convolution layers and residual blocks.
Specifically, we change the channel sizes of the last layers of the disparity map estimator and filter predictor blocks to $c_{e}$ so that they predict conventional feature maps.
We also replace the IAC layer with residual blocks that take a concatenation of $e_B$ and $\textbf{F}_{deblur}$ as input, and detach the reblurring network.
The baseline model is trained with neither \Loss{disp} nor \Loss{reblur} but only \Loss{deblur}.
For the other variants, we recover combinations of the restructured components one by one from the baseline model.
For the variants with the disparity map estimator, we train them also with \Loss{disp}.
Similarly, for the variants with the reblurring network, we train them also with \Loss{reblur}.
The detailed architectures of all the models used in the ablation study can be found in the supplementary material.
\begin{figure*}[t]
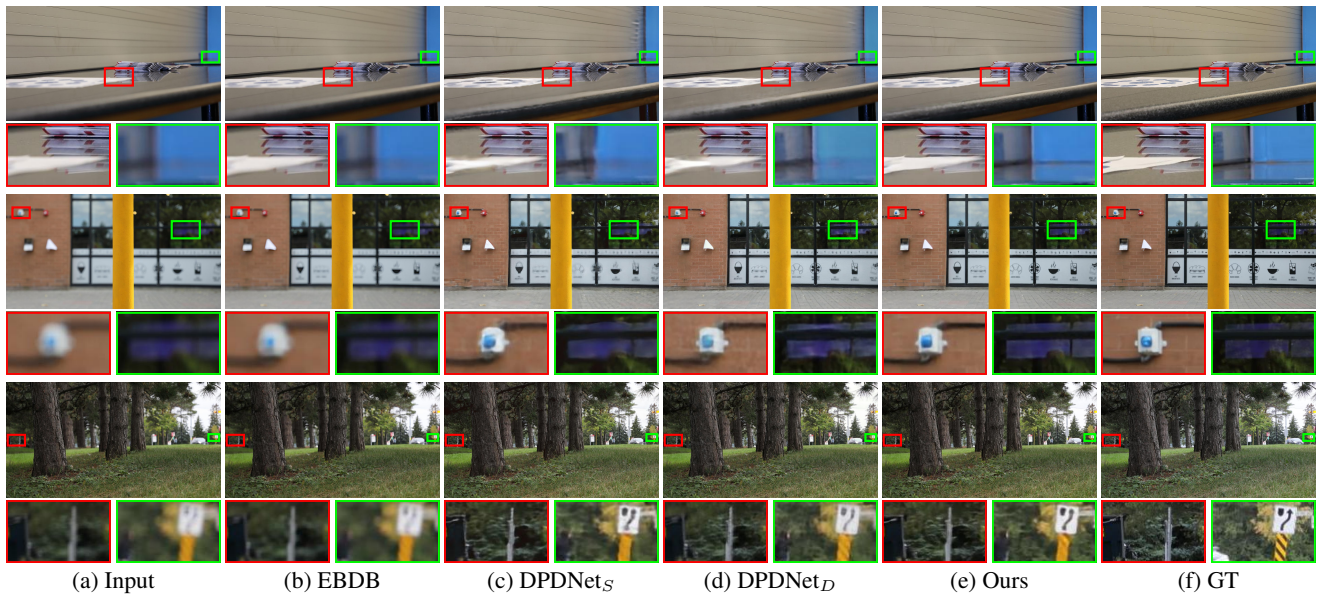

\centering
    \def \ext {.jpg} %
    \setlength\tabcolsep{0.8 pt}
    \def \wb {0.1635}
    \def \ws {0.07975}

    \def \corl {0.45} %
    \def \corb {0.30} %
    \def \corrt {0.15} %
    \def \corrty {0.16875} %
    
    \def \corls {0.9} %
    \def \corbs {0.5} %
    \def \corrts {0.1} %
    \def \corrtys {0.1125} %
    \def \img {62\ext} %

    \small
    \begin{tabular}{l@{\hspace{0pt}}rl@{\hspace{0pt}}rl@{\hspace{0pt}}rl@{\hspace{0pt}}rl@{\hspace{0pt}}rl@{\hspace{0pt}}r}
    
    \multicolumn{2}{c}{
    \begin{tikzpicture}
    \node[anchor=south west,inner sep=0] (image) at (0,0) {\includegraphics[trim=0 100px 0 100px, clip,  width=\wb\textwidth]{data/DPDD/figures/Input/\img}};
    \begin{scope}[x={(image.south east)},y={(image.north west)}]                                                 
        \draw[red,thick] ([shift={(0.01, 0.01)}]\corl, \corb) rectangle ([shift={(-0.01, -0.01)}]{\corl+\corrt}, {\corb+\corrty});
        \draw[green,thick] ([shift={(0.01, 0.01)}]\corls, \corbs) rectangle ([shift={(-0.01, -0.01)}]{\corls+\corrts}, {\corbs+\corrtys});
    \end{scope}
    \end{tikzpicture}} &
    
    \multicolumn{2}{c}{
    \begin{tikzpicture}
    \node[anchor=south west,inner sep=0] (image) at (0,0) {\includegraphics[trim=0 100px 0 100px, clip,  width=\wb\textwidth]{data/DPDD/figures/EBDB/\img}};
    \begin{scope}[x={(image.south east)},y={(image.north west)}]
        \draw[red,thick] ([shift={(0.01, 0.01)}]\corl, \corb) rectangle ([shift={(-0.01, -0.01)}]{\corl+\corrt}, {\corb+\corrty});
        \draw[green,thick] ([shift={(0.01, 0.01)}]\corls, \corbs) rectangle ([shift={(-0.01, -0.01)}]{\corls+\corrts}, {\corbs+\corrtys});
    \end{scope}
    \end{tikzpicture}} &
    
    \multicolumn{2}{c}{
    \begin{tikzpicture}
    \node[anchor=south west,inner sep=0] (image) at (0,0) {\includegraphics[trim=0 100px 0 100px, clip,  width=\wb\textwidth]{data/DPDD/figures/DPDNet/single/\img}};
    \begin{scope}[x={(image.south east)},y={(image.north west)}]
        \draw[red,thick] ([shift={(0.01, 0.01)}]\corl, \corb) rectangle ([shift={(-0.01, -0.01)}]{\corl+\corrt}, {\corb+\corrty});
        \draw[green,thick] ([shift={(0.01, 0.01)}]\corls, \corbs) rectangle ([shift={(-0.01, -0.01)}]{\corls+\corrts}, {\corbs+\corrtys});
    \end{scope}
    \end{tikzpicture}} &

    \multicolumn{2}{c}{
    \begin{tikzpicture}
    \node[anchor=south west,inner sep=0] (image) at (0,0) {\includegraphics[trim=0 100px 0 100px, clip,  width=\wb\textwidth]{data/DPDD/figures/DPDNet/dual/\img}};
    \begin{scope}[x={(image.south east)},y={(image.north west)}]
        \draw[red,thick] ([shift={(0.01, 0.01)}]\corl, \corb) rectangle ([shift={(-0.01, -0.01)}]{\corl+\corrt}, {\corb+\corrty});
        \draw[green,thick] ([shift={(0.01, 0.01)}]\corls, \corbs) rectangle ([shift={(-0.01, -0.01)}]{\corls+\corrts}, {\corbs+\corrtys});
    \end{scope}
    \end{tikzpicture}} &
    
    \multicolumn{2}{c}{
    \begin{tikzpicture}
    \node[anchor=south west,inner sep=0] (image) at (0,0) {\includegraphics[trim=0 100px 0 100px, clip,  width=\wb\textwidth]{data/DPDD/figures/Ours/single/\img}};
    \begin{scope}[x={(image.south east)},y={(image.north west)}]
        \draw[red,thick] ([shift={(0.01, 0.01)}]\corl, \corb) rectangle ([shift={(-0.01, -0.01)}]{\corl+\corrt}, {\corb+\corrty});
        \draw[green,thick] ([shift={(0.01, 0.01)}]\corls, \corbs) rectangle ([shift={(-0.01, -0.01)}]{\corls+\corrts}, {\corbs+\corrtys});
    \end{scope}
    \end{tikzpicture}} &
    
    \multicolumn{2}{c}{
    \begin{tikzpicture}
    \node[anchor=south west,inner sep=0] (image) at (0,0) {\includegraphics[trim=0 100px 0 100px, clip,  width=\wb\textwidth]{data/DPDD/figures/GT/\img}};
    \begin{scope}[x={(image.south east)},y={(image.north west)}]
        \draw[red,thick] ([shift={(0.01, 0.01)}]\corl, \corb) rectangle ([shift={(-0.01, -0.01)}]{\corl+\corrt}, {\corb+\corrty});
        \draw[green,thick] ([shift={(0.01, 0.01)}]\corls, \corbs) rectangle ([shift={(-0.01, -0.01)}]{\corls+\corrts}, {\corbs+\corrtys});
    \end{scope}
    \end{tikzpicture}} \\[-0.03in]
    
    \adjustbox{width=\ws\textwidth, trim={\corl\width} {\corb\height} {\width-(\corl\width+\corrt\width)} {\height-(\corb\height+\corrty\height)},clip,cfbox=red 0.7pt -0.7pt}
    {\includegraphics[trim=0 100px 0 100px, clip,  width=\ws\textwidth]{data/DPDD/figures/Input/\img}} &
    \adjustbox{width=\ws\textwidth, trim={\corls\width} {\corbs\height} {\width-(\corls\width+\corrts\width)} {\height-(\corbs\height+\corrtys\height)}, clip,cfbox=green 0.7pt -0.7pt}
    {\includegraphics[trim=0 100px 0 100px, clip,  width=\ws\textwidth]{data/DPDD/figures/Input/\img}} &    
    
    \adjustbox{width=\ws\textwidth, trim={\corl\width} {\corb\height} {\width-(\corl\width+\corrt\width)} {\height-(\corb\height+\corrty\height)},clip,cfbox=red 0.7pt -0.7pt}
    {\includegraphics[trim=0 100px 0 100px, clip,  width=\ws\textwidth]{data/DPDD/figures/EBDB/\img}} &
    \adjustbox{width=\ws\textwidth, trim={\corls\width} {\corbs\height} {\width-(\corls\width+\corrts\width)} {\height-(\corbs\height+\corrtys\height)}, clip,cfbox=green 0.7pt -0.7pt}
    {\includegraphics[trim=0 100px 0 100px, clip,  width=\ws\textwidth]{data/DPDD/figures/EBDB/\img}} &    
    
    \adjustbox{width=\ws\textwidth, trim={\corl\width} {\corb\height} {\width-(\corl\width+\corrt\width)} {\height-(\corb\height+\corrty\height)},clip,cfbox=red 0.7pt -0.7pt}
    {\includegraphics[trim=0 100px 0 100px, clip,  width=\ws\textwidth]{data/DPDD/figures/DPDNet/single/\img}} &
    \adjustbox{width=\ws\textwidth, trim={\corls\width} {\corbs\height} {\width-(\corls\width+\corrts\width)} {\height-(\corbs\height+\corrtys\height)}, clip,cfbox=green 0.7pt -0.7pt}
    {\includegraphics[trim=0 100px 0 100px, clip,  width=\ws\textwidth]{data/DPDD/figures/DPDNet/single/\img}} &    
    
    \adjustbox{width=\ws\textwidth, trim={\corl\width} {\corb\height} {\width-(\corl\width+\corrt\width)} {\height-(\corb\height+\corrty\height)},clip,cfbox=red 0.7pt -0.7pt}
    {\includegraphics[trim=0 100px 0 100px, clip,  width=\ws\textwidth]{data/DPDD/figures/DPDNet/dual/\img}} &
    \adjustbox{width=\ws\textwidth, trim={\corls\width} {\corbs\height} {\width-(\corls\width+\corrts\width)} {\height-(\corbs\height+\corrtys\height)}, clip,cfbox=green 0.7pt -0.7pt}
    {\includegraphics[trim=0 100px 0 100px, clip,  width=\ws\textwidth]{data/DPDD/figures/DPDNet/dual/\img}} &        
    
    \adjustbox{width=\ws\textwidth, trim={\corl\width} {\corb\height} {\width-(\corl\width+\corrt\width)} {\height-(\corb\height+\corrty\height)},clip,cfbox=red 0.7pt -0.7pt}
    {\includegraphics[trim=0 100px 0 100px, clip,  width=\ws\textwidth]{data/DPDD/figures/Ours/single/\img}} &
    \adjustbox{width=\ws\textwidth, trim={\corls\width} {\corbs\height} {\width-(\corls\width+\corrts\width)} {\height-(\corbs\height+\corrtys\height)}, clip,cfbox=green 0.7pt -0.7pt}
    {\includegraphics[trim=0 100px 0 100px, clip,  width=\ws\textwidth]{data/DPDD/figures/Ours/single/\img}} &       
    \adjustbox{width=\ws\textwidth, trim={\corl\width} {\corb\height} {\width-(\corl\width+\corrt\width)} {\height-(\corb\height+\corrty\height)},clip,cfbox=red 0.7pt -0.7pt}
    {\includegraphics[trim=0 100px 0 100px, clip,  width=\ws\textwidth]{data/DPDD/figures/GT/\img}} &
    \adjustbox{width=\ws\textwidth, trim={\corls\width} {\corbs\height} {\width-(\corls\width+\corrts\width)} {\height-(\corbs\height+\corrtys\height)}, clip,cfbox=green 0.7pt -0.7pt}
    {\includegraphics[trim=0 100px 0 100px, clip,  width=\ws\textwidth]{data/DPDD/figures/GT/\img}}     
    
    \\[-0.02in]
    \end{tabular}%
  
    \def \corl {0.022} %
    \def \corb {0.78} %
    \def \corrt {0.1} %
    \def \corrty {0.1125} %
    
    \def \corls {0.76} %
    \def \corbs {0.605} %
    \def \corrts {0.15} %
    \def \corrtys {0.16875} %
    \def \img {59\ext} %
    
    \small
    \begin{tabular}{l@{\hspace{0pt}}rl@{\hspace{0pt}}rl@{\hspace{0pt}}rl@{\hspace{0pt}}rl@{\hspace{0pt}}rl@{\hspace{0pt}}r}
    
    \multicolumn{2}{c}{
    \begin{tikzpicture}
    \node[anchor=south west,inner sep=0] (image) at (0,0) {\includegraphics[trim=0 100px 0 100px, clip,  width=\wb\textwidth]{data/DPDD/figures/Input/\img}};
    \begin{scope}[x={(image.south east)},y={(image.north west)}]
        \draw[red,thick] ([shift={(0.01, 0.01)}]\corl, \corb) rectangle ([shift={(-0.01, -0.01)}]{\corl+\corrt}, {\corb+\corrty});
        \draw[green,thick] ([shift={(0.01, 0.01)}]\corls, \corbs) rectangle ([shift={(-0.01, -0.01)}]{\corls+\corrts}, {\corbs+\corrtys});
    \end{scope}
    \end{tikzpicture}} &
    
    \multicolumn{2}{c}{
    \begin{tikzpicture}
    \node[anchor=south west,inner sep=0] (image) at (0,0) {\includegraphics[trim=0 100px 0 100px, clip,  width=\wb\textwidth]{data/DPDD/figures/EBDB/\img}};
    \begin{scope}[x={(image.south east)},y={(image.north west)}]
        \draw[red,thick] ([shift={(0.01, 0.01)}]\corl, \corb) rectangle ([shift={(-0.01, -0.01)}]{\corl+\corrt}, {\corb+\corrty});
        \draw[green,thick] ([shift={(0.01, 0.01)}]\corls, \corbs) rectangle ([shift={(-0.01, -0.01)}]{\corls+\corrts}, {\corbs+\corrtys});
    \end{scope}
    \end{tikzpicture}} &
    
    \multicolumn{2}{c}{
    \begin{tikzpicture}
    \node[anchor=south west,inner sep=0] (image) at (0,0) {\includegraphics[trim=0 100px 0 100px, clip,  width=\wb\textwidth]{data/DPDD/figures/DPDNet/single/\img}};
    \begin{scope}[x={(image.south east)},y={(image.north west)}]
        \draw[red,thick] ([shift={(0.01, 0.01)}]\corl, \corb) rectangle ([shift={(-0.01, -0.01)}]{\corl+\corrt}, {\corb+\corrty});
        \draw[green,thick] ([shift={(0.01, 0.01)}]\corls, \corbs) rectangle ([shift={(-0.01, -0.01)}]{\corls+\corrts}, {\corbs+\corrtys});
    \end{scope}
    \end{tikzpicture}} &

    \multicolumn{2}{c}{
    \begin{tikzpicture}
    \node[anchor=south west,inner sep=0] (image) at (0,0) {\includegraphics[trim=0 100px 0 100px, clip,  width=\wb\textwidth]{data/DPDD/figures/DPDNet/dual/\img}};
    \begin{scope}[x={(image.south east)},y={(image.north west)}]
        \draw[red,thick] ([shift={(0.01, 0.01)}]\corl, \corb) rectangle ([shift={(-0.01, -0.01)}]{\corl+\corrt}, {\corb+\corrty});
        \draw[green,thick] ([shift={(0.01, 0.01)}]\corls, \corbs) rectangle ([shift={(-0.01, -0.01)}]{\corls+\corrts}, {\corbs+\corrtys});
    \end{scope}
    \end{tikzpicture}} &
    
    \multicolumn{2}{c}{
    \begin{tikzpicture}
    \node[anchor=south west,inner sep=0] (image) at (0,0) {\includegraphics[trim=0 100px 0 100px, clip,  width=\wb\textwidth]{data/DPDD/figures/Ours/single/\img}};
    \begin{scope}[x={(image.south east)},y={(image.north west)}]
        \draw[red,thick] ([shift={(0.01, 0.01)}]\corl, \corb) rectangle ([shift={(-0.01, -0.01)}]{\corl+\corrt}, {\corb+\corrty});
        \draw[green,thick] ([shift={(0.01, 0.01)}]\corls, \corbs) rectangle ([shift={(-0.01, -0.01)}]{\corls+\corrts}, {\corbs+\corrtys});
    \end{scope}
    \end{tikzpicture}} &
    
    \multicolumn{2}{c}{
    \begin{tikzpicture}
    \node[anchor=south west,inner sep=0] (image) at (0,0) {\includegraphics[trim=0 100px 0 100px, clip,  width=\wb\textwidth]{data/DPDD/figures/GT/\img}};
    \begin{scope}[x={(image.south east)},y={(image.north west)}]
        \draw[red,thick] ([shift={(0.01, 0.01)}]\corl, \corb) rectangle ([shift={(-0.01, -0.01)}]{\corl+\corrt}, {\corb+\corrty});
        \draw[green,thick] ([shift={(0.01, 0.01)}]\corls, \corbs) rectangle ([shift={(-0.01, -0.01)}]{\corls+\corrts}, {\corbs+\corrtys});
    \end{scope}
    \end{tikzpicture}} \\[-0.03in]
    
    \adjustbox{width=\ws\textwidth, trim={\corl\width} {\corb\height} {\width-(\corl\width+\corrt\width)} {\height-(\corb\height+\corrty\height)},clip,cfbox=red 0.7pt -0.7pt}
    {\includegraphics[trim=0 100px 0 100px, clip,  width=\ws\textwidth]{data/DPDD/figures/Input/\img}} &
    \adjustbox{width=\ws\textwidth, trim={\corls\width} {\corbs\height} {\width-(\corls\width+\corrts\width)} {\height-(\corbs\height+\corrtys\height)}, clip,cfbox=green 0.7pt -0.7pt}
    {\includegraphics[trim=0 100px 0 100px, clip,  width=\ws\textwidth]{data/DPDD/figures/Input/\img}} &    
    
    \adjustbox{width=\ws\textwidth, trim={\corl\width} {\corb\height} {\width-(\corl\width+\corrt\width)} {\height-(\corb\height+\corrty\height)},clip,cfbox=red 0.7pt -0.7pt}
    {\includegraphics[trim=0 100px 0 100px, clip,  width=\ws\textwidth]{data/DPDD/figures/EBDB/\img}} &
    \adjustbox{width=\ws\textwidth, trim={\corls\width} {\corbs\height} {\width-(\corls\width+\corrts\width)} {\height-(\corbs\height+\corrtys\height)}, clip,cfbox=green 0.7pt -0.7pt}
    {\includegraphics[trim=0 100px 0 100px, clip,  width=\ws\textwidth]{data/DPDD/figures/EBDB/\img}} &    
    
    \adjustbox{width=\ws\textwidth, trim={\corl\width} {\corb\height} {\width-(\corl\width+\corrt\width)} {\height-(\corb\height+\corrty\height)},clip,cfbox=red 0.7pt -0.7pt}
    {\includegraphics[trim=0 100px 0 100px, clip,  width=\ws\textwidth]{data/DPDD/figures/DPDNet/single/\img}} &
    \adjustbox{width=\ws\textwidth, trim={\corls\width} {\corbs\height} {\width-(\corls\width+\corrts\width)} {\height-(\corbs\height+\corrtys\height)}, clip,cfbox=green 0.7pt -0.7pt}
    {\includegraphics[trim=0 100px 0 100px, clip,  width=\ws\textwidth]{data/DPDD/figures/DPDNet/single/\img}} &    
    
    \adjustbox{width=\ws\textwidth, trim={\corl\width} {\corb\height} {\width-(\corl\width+\corrt\width)} {\height-(\corb\height+\corrty\height)},clip,cfbox=red 0.7pt -0.7pt}
    {\includegraphics[trim=0 100px 0 100px, clip,  width=\ws\textwidth]{data/DPDD/figures/DPDNet/dual/\img}} &
    \adjustbox{width=\ws\textwidth, trim={\corls\width} {\corbs\height} {\width-(\corls\width+\corrts\width)} {\height-(\corbs\height+\corrtys\height)}, clip,cfbox=green 0.7pt -0.7pt}
    {\includegraphics[trim=0 100px 0 100px, clip,  width=\ws\textwidth]{data/DPDD/figures/DPDNet/dual/\img}} &        
    
    \adjustbox{width=\ws\textwidth, trim={\corl\width} {\corb\height} {\width-(\corl\width+\corrt\width)} {\height-(\corb\height+\corrty\height)},clip,cfbox=red 0.7pt -0.7pt}
    {\includegraphics[trim=0 100px 0 100px, clip,  width=\ws\textwidth]{data/DPDD/figures/Ours/single/\img}} &
    \adjustbox{width=\ws\textwidth, trim={\corls\width} {\corbs\height} {\width-(\corls\width+\corrts\width)} {\height-(\corbs\height+\corrtys\height)}, clip,cfbox=green 0.7pt -0.7pt}
    {\includegraphics[trim=0 100px 0 100px, clip,  width=\ws\textwidth]{data/DPDD/figures/Ours/single/\img}} &       
    \adjustbox{width=\ws\textwidth, trim={\corl\width} {\corb\height} {\width-(\corl\width+\corrt\width)} {\height-(\corb\height+\corrty\height)},clip,cfbox=red 0.7pt -0.7pt}
    {\includegraphics[trim=0 100px 0 100px, clip,  width=\ws\textwidth]{data/DPDD/figures/GT/\img}} &
    \adjustbox{width=\ws\textwidth, trim={\corls\width} {\corbs\height} {\width-(\corls\width+\corrts\width)} {\height-(\corbs\height+\corrtys\height)}, clip,cfbox=green 0.7pt -0.7pt}
    {\includegraphics[trim=0 100px 0 100px, clip,  width=\ws\textwidth]{data/DPDD/figures/GT/\img}}     
    
    \\[-0.02in]
    \end{tabular}%
    
    \def \corl {0.00} %
    \def \corb {0.44} %
    \def \corrt {0.1} %
    \def \corrty {0.1125} %
    
    \def \corls {0.93} %
    \def \corbs {0.48} %
    \def \corrts {0.07} %
    \def \corrtys {0.07875} %
    \def \img {36\ext} %
    
    \small
    \begin{tabular}{l@{\hspace{0pt}}rl@{\hspace{0pt}}rl@{\hspace{0pt}}rl@{\hspace{0pt}}rl@{\hspace{0pt}}rl@{\hspace{0pt}}r}
    
    \multicolumn{2}{c}{
    \begin{tikzpicture}
    \node[anchor=south west,inner sep=0] (image) at (0,0) {\includegraphics[trim=0 100px 0 100px, clip,  width=\wb\textwidth]{data/DPDD/figures/Input/\img}};
    \begin{scope}[x={(image.south east)},y={(image.north west)}]
        \draw[red,thick] ([shift={(0.01, 0.01)}]\corl, \corb) rectangle ([shift={(-0.01, -0.01)}]{\corl+\corrt}, {\corb+\corrty});
        \draw[green,thick] ([shift={(0.01, 0.01)}]\corls, \corbs) rectangle ([shift={(-0.01, -0.01)}]{\corls+\corrts}, {\corbs+\corrtys});
    \end{scope}
    \end{tikzpicture}} &
    
    \multicolumn{2}{c}{
    \begin{tikzpicture}
    \node[anchor=south west,inner sep=0] (image) at (0,0) {\includegraphics[trim=0 100px 0 100px, clip,  width=\wb\textwidth]{data/DPDD/figures/EBDB/\img}};
    \begin{scope}[x={(image.south east)},y={(image.north west)}]
        \draw[red,thick] ([shift={(0.01, 0.01)}]\corl, \corb) rectangle ([shift={(-0.01, -0.01)}]{\corl+\corrt}, {\corb+\corrty});
        \draw[green,thick] ([shift={(0.01, 0.01)}]\corls, \corbs) rectangle ([shift={(-0.01, -0.01)}]{\corls+\corrts}, {\corbs+\corrtys});
    \end{scope}
    \end{tikzpicture}} &
    
    \multicolumn{2}{c}{
    \begin{tikzpicture}
    \node[anchor=south west,inner sep=0] (image) at (0,0) {\includegraphics[trim=0 100px 0 100px, clip,  width=\wb\textwidth]{data/DPDD/figures/DPDNet/single/\img}};
    \begin{scope}[x={(image.south east)},y={(image.north west)}]
        \draw[red,thick] ([shift={(0.01, 0.01)}]\corl, \corb) rectangle ([shift={(-0.01, -0.01)}]{\corl+\corrt}, {\corb+\corrty});
        \draw[green,thick] ([shift={(0.01, 0.01)}]\corls, \corbs) rectangle ([shift={(-0.01, -0.01)}]{\corls+\corrts}, {\corbs+\corrtys});
    \end{scope}
    \end{tikzpicture}} &

    \multicolumn{2}{c}{
    \begin{tikzpicture}
    \node[anchor=south west,inner sep=0] (image) at (0,0) {\includegraphics[trim=0 100px 0 100px, clip,  width=\wb\textwidth]{data/DPDD/figures/DPDNet/dual/\img}};
    \begin{scope}[x={(image.south east)},y={(image.north west)}]
        \draw[red,thick] ([shift={(0.01, 0.01)}]\corl, \corb) rectangle ([shift={(-0.01, -0.01)}]{\corl+\corrt}, {\corb+\corrty});
        \draw[green,thick] ([shift={(0.01, 0.01)}]\corls, \corbs) rectangle ([shift={(-0.01, -0.01)}]{\corls+\corrts}, {\corbs+\corrtys});
    \end{scope}
    \end{tikzpicture}} &
    
    \multicolumn{2}{c}{
    \begin{tikzpicture}
    \node[anchor=south west,inner sep=0] (image) at (0,0) {\includegraphics[trim=0 100px 0 100px, clip,  width=\wb\textwidth]{data/DPDD/figures/Ours/single/\img}};
    \begin{scope}[x={(image.south east)},y={(image.north west)}]
        \draw[red,thick] ([shift={(0.01, 0.01)}]\corl, \corb) rectangle ([shift={(-0.01, -0.01)}]{\corl+\corrt}, {\corb+\corrty});
        \draw[green,thick] ([shift={(0.01, 0.01)}]\corls, \corbs) rectangle ([shift={(-0.01, -0.01)}]{\corls+\corrts}, {\corbs+\corrtys});
    \end{scope}
    \end{tikzpicture}} &
    
    \multicolumn{2}{c}{
    \begin{tikzpicture}
    \node[anchor=south west,inner sep=0] (image) at (0,0) {\includegraphics[trim=0 100px 0 100px, clip,  width=\wb\textwidth]{data/DPDD/figures/GT/\img}};
    \begin{scope}[x={(image.south east)},y={(image.north west)}]
        \draw[red,thick] ([shift={(0.01, 0.01)}]\corl, \corb) rectangle ([shift={(-0.01, -0.01)}]{\corl+\corrt}, {\corb+\corrty});
        \draw[green,thick] ([shift={(0.01, 0.01)}]\corls, \corbs) rectangle ([shift={(-0.01, -0.01)}]{\corls+\corrts}, {\corbs+\corrtys});
    \end{scope}
    \end{tikzpicture}} \\[-0.03in]
    
    \adjustbox{width=\ws\textwidth, trim={\corl\width} {\corb\height} {\width-(\corl\width+\corrt\width)} {\height-(\corb\height+\corrty\height)},clip,cfbox=red 0.7pt -0.7pt}
    {\includegraphics[trim=0 100px 0 100px, clip,  width=\ws\textwidth]{data/DPDD/figures/Input/\img}} &
    \adjustbox{width=\ws\textwidth, trim={\corls\width} {\corbs\height} {\width-(\corls\width+\corrts\width)} {\height-(\corbs\height+\corrtys\height)}, clip,cfbox=green 0.7pt -0.7pt}
    {\includegraphics[trim=0 100px 0 100px, clip,  width=\ws\textwidth]{data/DPDD/figures/Input/\img}} &    
    
    \adjustbox{width=\ws\textwidth, trim={\corl\width} {\corb\height} {\width-(\corl\width+\corrt\width)} {\height-(\corb\height+\corrty\height)},clip,cfbox=red 0.7pt -0.7pt}
    {\includegraphics[trim=0 100px 0 100px, clip,  width=\ws\textwidth]{data/DPDD/figures/EBDB/\img}} &
    \adjustbox{width=\ws\textwidth, trim={\corls\width} {\corbs\height} {\width-(\corls\width+\corrts\width)} {\height-(\corbs\height+\corrtys\height)}, clip,cfbox=green 0.7pt -0.7pt}
    {\includegraphics[trim=0 100px 0 100px, clip,  width=\ws\textwidth]{data/DPDD/figures/EBDB/\img}} &    
    
    \adjustbox{width=\ws\textwidth, trim={\corl\width} {\corb\height} {\width-(\corl\width+\corrt\width)} {\height-(\corb\height+\corrty\height)},clip,cfbox=red 0.7pt -0.7pt}
    {\includegraphics[trim=0 100px 0 100px, clip,  width=\ws\textwidth]{data/DPDD/figures/DPDNet/single/\img}} &
    \adjustbox{width=\ws\textwidth, trim={\corls\width} {\corbs\height} {\width-(\corls\width+\corrts\width)} {\height-(\corbs\height+\corrtys\height)}, clip,cfbox=green 0.7pt -0.7pt}
    {\includegraphics[trim=0 100px 0 100px, clip,  width=\ws\textwidth]{data/DPDD/figures/DPDNet/single/\img}} &    
    
    \adjustbox{width=\ws\textwidth, trim={\corl\width} {\corb\height} {\width-(\corl\width+\corrt\width)} {\height-(\corb\height+\corrty\height)},clip,cfbox=red 0.7pt -0.7pt}
    {\includegraphics[trim=0 100px 0 100px, clip,  width=\ws\textwidth]{data/DPDD/figures/DPDNet/dual/\img}} &
    \adjustbox{width=\ws\textwidth, trim={\corls\width} {\corbs\height} {\width-(\corls\width+\corrts\width)} {\height-(\corbs\height+\corrtys\height)}, clip,cfbox=green 0.7pt -0.7pt}
    {\includegraphics[trim=0 100px 0 100px, clip,  width=\ws\textwidth]{data/DPDD/figures/DPDNet/dual/\img}} &        
    
    \adjustbox{width=\ws\textwidth, trim={\corl\width} {\corb\height} {\width-(\corl\width+\corrt\width)} {\height-(\corb\height+\corrty\height)},clip,cfbox=red 0.7pt -0.7pt}
    {\includegraphics[trim=0 100px 0 100px, clip,  width=\ws\textwidth]{data/DPDD/figures/Ours/single/\img}} &
    \adjustbox{width=\ws\textwidth, trim={\corls\width} {\corbs\height} {\width-(\corls\width+\corrts\width)} {\height-(\corbs\height+\corrtys\height)}, clip,cfbox=green 0.7pt -0.7pt}
    {\includegraphics[trim=0 100px 0 100px, clip,  width=\ws\textwidth]{data/DPDD/figures/Ours/single/\img}} &       
    \adjustbox{width=\ws\textwidth, trim={\corl\width} {\corb\height} {\width-(\corl\width+\corrt\width)} {\height-(\corb\height+\corrty\height)},clip,cfbox=red 0.7pt -0.7pt}
    {\includegraphics[trim=0 100px 0 100px, clip,  width=\ws\textwidth]{data/DPDD/figures/GT/\img}} &
    \adjustbox{width=\ws\textwidth, trim={\corls\width} {\corbs\height} {\width-(\corls\width+\corrts\width)} {\height-(\corbs\height+\corrtys\height)}, clip,cfbox=green 0.7pt -0.7pt}
    {\includegraphics[trim=0 100px 0 100px, clip,  width=\ws\textwidth]{data/DPDD/figures/GT/\img}}     
    
    \\[-0.02in]
    
    \multicolumn{2}{c}{(a) Input} &
    \multicolumn{2}{c}{(b) EBDB} &
    \multicolumn{2}{c}{(c) DPDNet$_{S}$} &
    \multicolumn{2}{c}{(d) DPDNet$_{D}$} &
    \multicolumn{2}{c}{(e) Ours} &
    \multicolumn{2}{c}{(f) GT}
    \\
  \end{tabular}%
  
  \vspace{0.05cm}
 \caption{Qualitative comparison on the DPDD dataset \cite{Abuolaim:2020:DPDNet}.
The first and last columns show defocused input images and their ground-truth all-in-focus images, respectively.
Between the columns, we show deblurring results of different methods. %
Note that DPDNet$_{D}$ requires a pair of dual-pixel stereo images as input, while other methods, including ours, require only a single image at test time. 
Refer to the supplementary material for more results.
}
  \label{fig:comparison:DP}
\vspace{-14pt}
\end{figure*}

\vspace{-13pt}
\paragraph{Explicit deblurring filter prediction}
We first verify the effect of the filter prediction scheme implemented using the filter predictor and IAC layer.
\Tbl{\ref{tbl:ablation}} shows that
introducing the filter predictor and IAC layer to the baseline model increases the deblurring performance (the first and third rows in the table),
confirming the advantage of \textit{explicit} pixel-wise filter prediction in flexible handling of spatially-varying and large defocus blur.
In addition,
compared to the gain (PSNR: $0.36\pct$ and LPIPS: $3.21\pct$) obtained when the disparity map estimator is embedded in the baseline model (the second row in the table),
there is more performance gain (PSNR: $0.44\pct$ and LPIPS: $16.31\pct$) when the disparity map estimator is added to a model with the filter predictor and IAC layer (the fourth row in the table).
This observation validates that explicit utilization of deblurring filters has more potential in absorbing extra defocus blur-specific supervision provided by dual-pixel images.
\Fig{\ref{fig:ablation}} shows a qualitative comparison.
As shown in the figure, the filter predictor and IAC layer substantially enhance the deblurring quality ((b) and (c) vs. (d) and (e) in the figure).

\vspace{-13pt}
\paragraph{Disparity map estimation and reblurring}
We analyze the influence of the disparity map estimator and reblurring network.
Specifically, we compare the combinations $(\text{filter predictor}\,\texttt{+}\,\textrm{IAC}\,\texttt{+}\,\textrm{disparity map estimator)}$ and $(\text{filter predictor}\,\texttt{+}\,\textrm{IAC}\,\texttt{+}\,\textrm{reblurring network})$.
\Tbl{\ref{tbl:ablation}} show that the model with the disparity map estimator performs better than the model with the reblurring network in recovering textures (lower LPIPS), as the disparity map estimator helps more accurately estimate per-pixel blur amounts (\Fig{\ref{fig:ablation}}e and f).
On the other hand, the model with the reblurring network better restores overall image contents (higher PSNR), as the reblurring network guides deblurring filters to contain information about blur shapes and blur amounts.
We can also observe that the model with both disparity map estimator and reblurring network achieves the best performance in every measure (\Fig{\ref{fig:ablation}}g).
This shows that the disparity map estimator and reblurring network have synergistic effects, complementing each other. %

\subsection{Comparison with Previous Methods}
In this section, we compare our method with previous defocus map-based approaches as well as recent end-to-end learning-based approaches:
Just Noticeable Blur estimation (JNB) \cite{Shi:2015:JNB}, 
Edge-Based Defocus Blur estimation (EBDB) \cite{Karaali:2018:DMEAdaptive},
Defocus Map Estimation Network (DMENet) \cite{Lee:2019:DMENet},
DPDNet$_{S}$ and DPDNet$_{D}$ \cite{Abuolaim:2020:DPDNet}.
Among these, JNB, EBDB, and DMENet are defocus map-based approaches that first estimate a defocus map and perform non-blind deconvolution.
DPDNet$_{S}$ and DPDNet$_{D}$ are end-to-end learning-based approaches that directly restore all-in-focus images.
They share the same network architecture, but DPDNet$_{S}$ takes a single defocused image as input while DPDNet$_{D}$ takes a pair of dual-pixel stereo images.

For all the previous methods, we use the code (and model weights for learning-based methods, DMENet and DPDNet) provided by the authors.
For JNB, EBDB, and DMENet, we used the deconvolution method \cite{Krishnan:2008:deconvolution} to obtain all-in-focus images using the estimated defocus maps.
For DPDNet$_{S}$, we retrain the network on 8-bit images with the training code provided by the author, as the authors provide only a model trained with 16-bit images (refer to the supplementary material for the evaluation of our model trained on 16-bit images).
We measure computational costs in terms of the number of network parameters, number of multiply-accumulate operations (MACs) computed on a $1280\!\times\!720$ image, and average computation time computed on test images. %
For JNB and EBDB, which are not learning-based methods, we measure only their computation times.

We compare the methods on the DPDD test set \cite{Abuolaim:2020:DPDNet}. 
\Tbl{\ref{tbl:comparison:DP}} shows a quantitative comparison.
The previous defocus map-based methods show poor performance on the real-world blurred images in the DPDD test set, which is even lower than the input defocused images due to their restrictive blur models.
On the other hand, the recent end-to-end approaches, DPDNet$_{D}$ and DPDNet$_{S}$, achieve higher quality compared to the previous methods.
Our model outperforms DPDNet$_{S}$ by a significant gap with a smaller computational cost.
Moreover, although our model uses a single defocused image, it outperforms DPDNet$_{D}$ as well, proving the effectiveness of our approach.

\Fig{\ref{fig:comparison:DP}} shows a qualitative comparison.
Due to their inaccurate defocus maps and restricted blur models, the results of the defocus map-based methods have a large amount of remaining blur (\Fig{\ref{fig:comparison:DP}}b).
DPDNet$_{S}$ and DPDNet$_{D}$ produce better results than the previous ones, however, tend to produce artifacts and remaining blur (\Fig{\ref{fig:comparison:DP}}c and \ref{fig:comparison:DP}d).
On the other hand, our method shows more accurate deblurring results (\Fig{\ref{fig:comparison:DP}}e), even with a single defocused input.
Especially, compared to DPDNet$_{D}$, \Fig{\ref{fig:comparison:DP}} show that our method better handles spatially-varying blur (the first row), large blur (the second row), and image structures as well as textures (the third row).
More qualitative results are in the supplementary material.

\vspace{-10pt}
\subsubsection{Generalization Ability}
\label{ssec:comp:Real}

\vspace{-3pt}
As our method is trained using the DPDD training set, which is captured by a specific camera (Canon EOS 5D Mark IV), one question naturally follows how well the model generalizes to other images from different cameras.
To answer this question, we evaluate the performance of our approach on other test sets.
\begin{table}[t]
\centering
\small
\setlength\tabcolsep{0pt}
\begin{tabularx}{\columnwidth}{@{\hspace{5pt}}lYYYY}
\toprule
Model & PSNR$\uparrow$\! & SSIM$\uparrow$\! & MAE\MAES$\downarrow$\! & LPIPS$\downarrow$\! \\
\midrule
Input 
& 22.33 & 0.633 & 0.513 & 0.524 \\
\arrayrulecolor{lightgray}\cmidrule{1-5}
JNB \cite{Shi:2015:JNB}
& 22.36 & 0.635 & 0.511 & 0.601 \\ %

EBDB \cite{Karaali:2018:DMEAdaptive}
& 22.38 & 0.638 & 0.509 & 0.594 \\

DMENet \cite{Lee:2019:DMENet}
& 22.41 & 0.639 & 0.508 & 0.597 \\ %
DPDNet$_{S}$ \cite{Abuolaim:2020:DPDNet}
& 22.67 & 0.666 & 0.506 & 0.420 \\ %

\arrayrulecolor{lightgray}\cmidrule{1-5}
Ours
& \textbf{24.71} & \textbf{0.748} & \textbf{0.407} & \textbf{0.306} \\
\arrayrulecolor{black}\bottomrule

\end{tabularx}

\vspace{0.1cm}
\caption{Quantitative evaluation on the RealDOF test set.}
\vspace{-5pt}
\label{tbl:comparison:Real}
\end{table}

\begin{figure}[t]
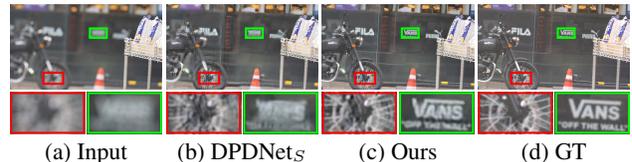

    \centering
    \def \ext {.jpg} %

    \def \corl {0.21} %
    \def \corb {0.06} %
    \def \corrt {0.15} %
    \def \corrty {0.15} %
    
    \def \corls {0.515} %
    \def \corbs {0.599} %
    \def \corrts {0.13} %
    \def \corrtys {0.13} %
    \def \img {010\ext}

    \setlength\tabcolsep{0.8 pt}
    \def \wb {0.242}
    \def \ws {0.119}
    
    \small
    \begin{tabular}{l@{\hspace{0pt}}rl@{\hspace{0pt}}rl@{\hspace{0pt}}rl@{\hspace{0pt}}r}
    
    \multicolumn{2}{c}{
    \begin{tikzpicture}
    \node[anchor=south west,inner sep=0] (image) at (0,0) {\includegraphics[trim=550px 0 20px 440px, clip,  width=\wb\columnwidth]{data/experiments/comparison_Real/figures/Input/\img}};
    \begin{scope}[x={(image.south east)},y={(image.north west)}]
        \draw[red,thick] ([shift={(0.01, 0.01)}]\corl, \corb) rectangle ([shift={(-0.01, -0.01)}]{\corl+\corrt}, {\corb+\corrty});
        \draw[green,thick] ([shift={(0.01, 0.01)}]\corls, \corbs) rectangle ([shift={(-0.01, -0.01)}]{\corls+\corrts}, {\corbs+\corrtys});
    \end{scope}
    \end{tikzpicture}} &
    
    \multicolumn{2}{c}{
    \begin{tikzpicture}
    \node[anchor=south west,inner sep=0] (image) at (0,0) {\includegraphics[trim=550px 0 20px 440px, clip,  width=\wb\columnwidth]{data/experiments/comparison_Real/figures/DPDNet/\img}};
    \begin{scope}[x={(image.south east)},y={(image.north west)}]
        \draw[red,thick] ([shift={(0.01, 0.01)}]\corl, \corb) rectangle ([shift={(-0.01, -0.01)}]{\corl+\corrt}, {\corb+\corrty});
        \draw[green,thick] ([shift={(0.01, 0.01)}]\corls, \corbs) rectangle ([shift={(-0.01, -0.01)}]{\corls+\corrts}, {\corbs+\corrtys});
    \end{scope}
    \end{tikzpicture}} &
    
    \multicolumn{2}{c}{
    \begin{tikzpicture}
    \node[anchor=south west,inner sep=0] (image) at (0,0) {\includegraphics[trim=550px 0 20px 440px, clip,  width=\wb\columnwidth]{data/experiments/comparison_Real/figures/Ours/\img}};
    \begin{scope}[x={(image.south east)},y={(image.north west)}]
        \draw[red,thick] ([shift={(0.01, 0.01)}]\corl, \corb) rectangle ([shift={(-0.01, -0.01)}]{\corl+\corrt}, {\corb+\corrty});
        \draw[green,thick] ([shift={(0.01, 0.01)}]\corls, \corbs) rectangle ([shift={(-0.01, -0.01)}]{\corls+\corrts}, {\corbs+\corrtys});
    \end{scope}
    \end{tikzpicture}} &
    
    \multicolumn{2}{c}{
    \begin{tikzpicture}
    \node[anchor=south west,inner sep=0] (image) at (0,0) {\includegraphics[trim=550px 0 20px 440px, clip,  width=\wb\columnwidth]{data/experiments/comparison_Real/figures/GT/\img}};
    \begin{scope}[x={(image.south east)},y={(image.north west)}]
        \draw[red,thick] ([shift={(0.01, 0.01)}]\corl, \corb) rectangle ([shift={(-0.01, -0.01)}]{\corl+\corrt}, {\corb+\corrty});
        \draw[green,thick] ([shift={(0.01, 0.01)}]\corls, \corbs) rectangle ([shift={(-0.01, -0.01)}]{\corls+\corrts}, {\corbs+\corrtys});
    \end{scope}
    \end{tikzpicture}}
    
    \\[-0.03in]
    
    \adjustbox{width=\ws\columnwidth, trim={\corl\width} {\corb\height} {\width-(\corl\width+\corrt\width)} {\height-(\corb\height+\corrty\height)},clip,cfbox=red 0.7pt -0.7pt}
    {\includegraphics[trim=550px 0 20px 440px, clip,  width=\ws\columnwidth]{data/experiments/comparison_Real/figures/Input/\img}} &
    \adjustbox{width=\ws\columnwidth, trim={\corls\width} {\corbs\height} {\width-(\corls\width+\corrts\width)} {\height-(\corbs\height+\corrtys\height)}, clip,cfbox=green 0.7pt -0.7pt}
    {\includegraphics[trim=550px 0 20px 440px, clip,  width=\ws\columnwidth]{data/experiments/comparison_Real/figures/Input/\img}} &

    \adjustbox{width=\ws\columnwidth, trim={\corl\width} {\corb\height} {\width-(\corl\width+\corrt\width)} {\height-(\corb\height+\corrty\height)},clip,cfbox=red 0.7pt -0.7pt}
    {\includegraphics[trim=550px 0 20px 440px, clip,  width=\ws\columnwidth]{data/experiments/comparison_Real/figures/DPDNet/\img}} &
    \adjustbox{width=\ws\columnwidth, trim={\corls\width} {\corbs\height} {\width-(\corls\width+\corrts\width)} {\height-(\corbs\height+\corrtys\height)}, clip,cfbox=green 0.7pt -0.7pt}
    {\includegraphics[trim=550px 0 20px 440px, clip,  width=\ws\columnwidth]{data/experiments/comparison_Real/figures/DPDNet/\img}} &
    
    \adjustbox{width=\ws\columnwidth, trim={\corl\width} {\corb\height} {\width-(\corl\width+\corrt\width)} {\height-(\corb\height+\corrty\height)},clip,cfbox=red 0.7pt -0.7pt}
    {\includegraphics[trim=550px 0 20px 440px, clip,  width=\ws\columnwidth]{data/experiments/comparison_Real/figures/Ours/\img}} &
    \adjustbox{width=\ws\columnwidth, trim={\corls\width} {\corbs\height} {\width-(\corls\width+\corrts\width)} {\height-(\corbs\height+\corrtys\height)}, clip,cfbox=green 0.7pt -0.7pt}
    {\includegraphics[trim=550px 0 20px 440px, clip,  width=\ws\columnwidth]{data/experiments/comparison_Real/figures/Ours/\img}} &

    \adjustbox{width=\ws\columnwidth, trim={\corl\width} {\corb\height} {\width-(\corl\width+\corrt\width)} {\height-(\corb\height+\corrty\height)}, clip,cfbox=red 0.7pt -0.7pt}
    {\includegraphics[trim=550px 0 20px 440px, clip,  width=\ws\columnwidth]{data/experiments/comparison_Real/figures/GT/\img}} &
    \adjustbox{width=\ws\columnwidth, trim={\corls\width} {\corbs\height} {\width-(\corls\width+\corrts\width)} {\height-(\corbs\height+\corrtys\height)}, clip,cfbox=green 0.7pt -0.7pt}
    {\includegraphics[trim=550px 0 20px 440px, clip,  width=\ws\columnwidth]{data/experiments/comparison_Real/figures/GT/\img}}
    
    \\[-0.02in]
    \multicolumn{2}{c}{(a) Input} &
    \multicolumn{2}{c}{(b) DPDNet$_{S}$} &
    \multicolumn{2}{c}{(c) Ours} &
    \multicolumn{2}{c}{(d) GT}
  \end{tabular}
  
 \vspace{0.05cm}
 \caption{Qualitative comparison on the RealDOF test set.
From left to right: a defocused input image, deblurred results of DPDNet$_{S}$ \cite{Abuolaim:2020:DPDNet} and our method, and a ground-truth image.
}
  \vspace{-12pt}
  \label{fig:comparison:Real}
\end{figure}

\vspace{-13pt}
\paragraph{RealDOF test set}
To quantitatively measure the performance of our method on real-world defocus blur images, we prepare a new dataset named \textit{Real Depth of Field (RealDOF)} test set.
RealDOF consists of 50 scenes. For each scene, the dataset provides a pair of a defocused image and its corresponding all-in-focus image.
To capture image pairs of the same scene with different depth-of-fields, we built a dual-camera system with a beam splitter as described in \cite{rim:2020:ECCV}.
Specifically, our system is equipped with two cameras attached to the vertical rig with a beam splitter.
We used two Sony a7R IV cameras, which do not support dual pixels, with Sony 135mm F1.8 lenses.
The system is also equipped with a multi-camera trigger to synchronize the camera shutters to capture images simultaneously.
The captured images are post-processed for geometric and photometric alignments, similarly to \cite{rim:2020:ECCV}.
More details about the RealDOF test set are in the supplementary material.

\Tbl{\ref{tbl:comparison:Real}} shows a quantitative comparison on the RealDOF test set.
The table shows that our model clearly improves the image quality, showing that the model can generalize well to images from other cameras.
Moreover, our model significantly outperforms the previous state-of-the-art single image deblurring method, DPDNet$_{S}$, by more than 2\,dB in terms of PSNR.
\Fig{\ref{fig:comparison:Real}} qualitatively compares our method and DPDNet$_{S}$.
While the result of DPDNet$_{S}$ contains some amount of remaining blur, ours looks much sharper with no remaining blur.

\vspace{-13pt}
\paragraph{CUHK blur detection dataset}
The CUHK blur detection dataset \cite{Shi:2014:CUHK} provides 704 defocused images collected from the internet without ground-truth all-in-focus images.
\Fig{\ref{fig:comparison:CUHK}} shows a qualitative comparison between DPDNet$_{S}$ and ours on the CUHK dataset.
The result shows that our method removes defocus blur and restores fine details more successfully than DPDNet$_{S}$.
Refer to the supplementary material for more qualitative results.

\begin{table}[t]
\centering

\small
\setlength{\tabcolsep}{0pt}
\begin{tabularx}{\linewidth}{
    @{\hspace{2pt}}
    cc
    Y
    Y
    Y
    Y
    c
    @{\hspace{3pt}}
    c
    @{\hspace{2pt}}
}
\toprule
\multirow{2}{*}[-0.5\dimexpr \aboverulesep + \belowrulesep + \cmidrulewidth]{$N$}
& \multirow{2}{*}[-0.5\dimexpr \aboverulesep + \belowrulesep + \cmidrulewidth]{(RF)}
& \multicolumn{4}{c}{Deblurring Results}
& \multirow{2}{*}[-0.5\dimexpr \aboverulesep + \belowrulesep + \cmidrulewidth]{\makecell{Params\\(M)}}
& \multirow{2}{*}[-0.5\dimexpr \aboverulesep + \belowrulesep + \cmidrulewidth]{\makecell{MACs\\(B)}}
\\

\cmidrule(lr){3-6}
& & PSNR$\uparrow$\! & SSIM$\uparrow$\! & MAE\MAES$\downarrow$\! & LPIPS$\downarrow$\!& &
\\
\midrule
8 & (17)
& 25.19 & 0.777 & 0.404 & 0.246 & \textbf{9.44}  & \textbf{347.9} \\

17 & (35)
& 25.37 & 0.789 & 0.394 & 0.217 & 10.48 & 362.9  \\ %

26 & (53)
& 25.39 & 0.788 & 0.393 & 0.215 & 11.52 & 377.9 \\ %

35 & (71)
& 25.42 & 0.789 & 0.391 & 0.213 & 12.56 & 392.8 \\ %

44 & (89)
& \textbf{25.45} & \textbf{0.792} & \textbf{0.389} & \textbf{0.206} & 13.60 & 407.8 \\ %

\bottomrule
\end{tabularx}

\vspace{0.1cm}
\caption{Deblurring performance and computational cost with respect to the number of deblurring filters $N$ evaluated on the DPDD dataset \cite{Abuolaim:2020:DPDNet}. RF denotes the receptive field size.}
\vspace{-5pt}
\label{tbl:effect:stack}
\end{table}
\begin{figure}[t]
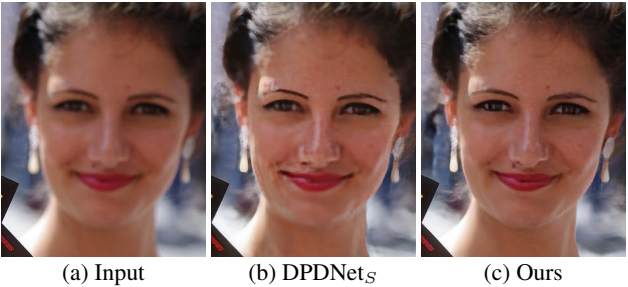

    \centering
    \def \ext {.jpg} %

    \def \corl {0.22} %
    \def \corb {0.8} %
    \def \corrt {0.15} %
    \def \corrty {0.15} %
    
    \def \corls {0.78} %
    \def \corbs {0.10} %
    \def \corrts {0.15} %
    \def \corrtys {0.15} %
    \def \img {666\ext}

    \setlength\tabcolsep{0.0 pt}
    \def \wb {0.325}
    \def \ws {0.145}
    
    \small
    \begin{tabularx}{\linewidth}{@{}Y@{\hspace{1pt}}Y@{\hspace{1pt}}Y@{}}
    
    \begin{tikzpicture}
    \node[anchor=south west,inner sep=0] (image) at (0,0) {\includegraphics[trim=330px 30px 0 0, clip,  width=\wb\columnwidth]{data/experiments/comparison_CUHK/figures/Input/\img}};
    \begin{scope}[x={(image.south east)},y={(image.north west)}]
    \end{scope}
    \end{tikzpicture} &
    
    \begin{tikzpicture}
    \node[anchor=south west,inner sep=0] (image) at (0,0) {\includegraphics[trim=330px 30px 0 0, clip,  width=\wb\columnwidth]{data/experiments/comparison_CUHK/figures/DPDNet/\img}};
    \begin{scope}[x={(image.south east)},y={(image.north west)}]
    \end{scope}
    \end{tikzpicture} &
    
    \begin{tikzpicture}
    \node[anchor=south west,inner sep=0] (image) at (0,0) {\includegraphics[trim=330px 30px 0 0, clip,  width=\wb\columnwidth]{data/experiments/comparison_CUHK/figures/Ours/\img}};
    \begin{scope}[x={(image.south east)},y={(image.north west)}]
    \end{scope}
    \end{tikzpicture} 
    
    \\
    [-0.18in]
    (a) Input & (b) DPDNet$_{S}$ & (c) Ours
  \end{tabularx}%
  
 \vspace{0.05cm}  
 \caption{Qualitative comparison on the CUHK blur detection dataset \cite{Shi:2014:CUHK}.
From left to right: a defocused input image, deblurred results of DPDNet$_{S}$ \cite{Abuolaim:2020:DPDNet} and our method.
}
  \vspace{-12pt}
  \label{fig:comparison:CUHK}
\end{figure}

\vspace{-13pt}
\paragraph{Pixel dual-pixel test set}
The DPDD dataset \cite{Abuolaim:2020:DPDNet} provide an additional test set consisting of dual-pixel defocused images captured by a Google Pixel 4 smartphone camera.
We refer the readers to the supplementary material for the qualitative evaluation on the Pixel dual-pixel test set.

\subsection{Analysis on IFAN}
In this section, we further investigate the effect of different components of IFAN.
We first analyze the effect of the number of separable filters $N$ in $\textbf{F}_{deblur}$.
Then, we evaluate the effect of IAC compared to \Net{FAC} \cite{Zhou:2019:STFAN}.

\vspace{-13pt}
\paragraph{Number of separable filters $N$}
A larger number of separable filters in $\textbf{F}_{deblur}$ leads IFAN to establish larger receptive fields and more accurate deblurring filters.
Consequently, 
when $N$ is larger,
we can handle large defocus blur more accurately.
\Tbl{\ref{tbl:effect:stack}} compares the performances with different values of $N$,
where
the deblurring quality increases with $N$.
Based on the result, we choose $N\,\texttt{=}\,17$ for our final model, as the improvement is small for $N\,\texttt{>}\,17$.

\vspace{-13pt}
\paragraph{IAC vs. FAC}
Finally, we analyze the effectiveness of the proposed IAC compared to FAC \cite{Zhou:2019:STFAN}.
To compare IAC and FAC, we replace the IAC layers in both IFAN and reblurring network in our final model with FAC layers.
For the FAC layers, we use $k~\texttt{=}\,11$ to match the computational cost to that of our final model for fairness.
\Tbl{\ref{tbl:effect:IAC}} and \Fig{\ref{fig:effect:IAC}} respectively show quantitative and qualitative comparisons between IAC and FAC.
The comparisons show that IAC outperforms FAC even with fewer parameters and operations, as IAC is better in handling large defocus blur by covering a much larger receptive field ($35\!\times\!35$) on defocused features than the receptive field ($11\!\times\!11$) of FAC.

\begin{table}[t]{
\centering
\small
\setlength{\tabcolsep}{0pt}
\begin{tabularx}{\linewidth}{
    @{\hspace{1pt}}
    c
    Y
    Y
    Y
    Y
    c
    @{\hspace{2pt}}
    c
    @{\hspace{1pt}}
}
\toprule

 \multirow{2}{*}[-0.5\dimexpr \aboverulesep + \belowrulesep + \cmidrulewidth]{Module}
& \multicolumn{4}{c}{Deblurring Results}
& \multirow{2}{*}[-0.5\dimexpr \aboverulesep + \belowrulesep + \cmidrulewidth]{\makecell{Params\\(M)}}
& \multirow{2}{*}[-0.5\dimexpr \aboverulesep + \belowrulesep + \cmidrulewidth]{\makecell{MACs\\(B)}}
\\
\cmidrule(lr){2-5}
 & PSNR$\uparrow$ & SSIM$\uparrow$ & MAE\MAES$\downarrow$ & LPIPS$\downarrow$ & &
\\
\midrule
FAC %
& 25.18 & 0.778 & 0.406 & 0.249 & 10.51 & 363.4 \\

\Net{IAC}
& \textbf{25.37} & \textbf{0.789} & \textbf{0.394} & \textbf{0.217} & \textbf{10.48} & \textbf{362.9}  \\[-0.02in]

\bottomrule
\end{tabularx}

\vspace{0.1cm}
\caption{Quantitative comparison between the FAC \cite{Zhou:2019:STFAN} and IAC layers evaluated on the DPDD dataset \cite{Abuolaim:2020:DPDNet}.
IAC indicates our model with IAC layers, while FAC indicates a variant of our model whose IAC layers are replaced with FAC layers.
We set the filter size to $11\!\times\!11$ for the FAC layers for fairness in computational costs.
}
\vspace{-5pt}
\label{tbl:effect:IAC}
}
\end{table}
\begin{figure}[t]
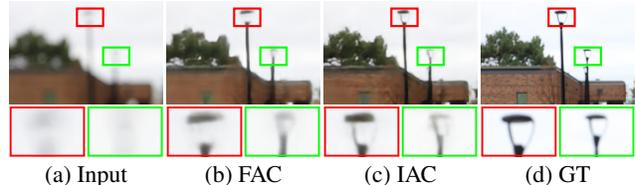

    \centering
    \def \ext {.jpg} %

    \def \corl {0.43} %
    \def \corb {0.748} %
    \def \corrt {0.19} %
    \def \corrty {0.19} %
    
    \def \corls {0.60} %
    \def \corbs {0.37} %
    \def \corrts {0.19} %
    \def \corrtys {0.19} %
    \def \img {35\ext}
    
    \setlength\tabcolsep{0.8 pt}
    \def \wb {0.244}
    \def \ws {0.120}
    
    \small
    \begin{tabular}{l@{\hspace{0pt}}rl@{\hspace{0pt}}rl@{\hspace{0pt}}rl@{\hspace{0pt}}r}
    
    \multicolumn{2}{c}{
    \begin{tikzpicture}
    \node[anchor=south west,inner sep=0] (image) at (0,0) {\includegraphics[trim=0 500px 900px 100px, clip,  width=\wb\columnwidth]{data/DPDD/figures/Input/\img}};
    \begin{scope}[x={(image.south east)},y={(image.north west)}]
        \draw[red,thick] ([shift={(0.01, 0.01)}]\corl, \corb) rectangle ([shift={(-0.01, -0.01)}]{\corl+\corrt}, {\corb+\corrty});
        \draw[green,thick] ([shift={(0.01, 0.01)}]\corls, \corbs) rectangle ([shift={(-0.01, -0.01)}]{\corls+\corrts}, {\corbs+\corrtys});
    \end{scope}
    \end{tikzpicture}} &
    
    \multicolumn{2}{c}{
    \begin{tikzpicture}
    \node[anchor=south west,inner sep=0] (image) at (0,0) {\includegraphics[trim=0 500px 900px 100px, clip,  width=\wb\columnwidth]{data/experiments/effect/figures/FAC/\img}};
    \begin{scope}[x={(image.south east)},y={(image.north west)}]
        \draw[red,thick] ([shift={(0.01, 0.01)}]\corl, \corb) rectangle ([shift={(-0.01, -0.01)}]{\corl+\corrt}, {\corb+\corrty});
        \draw[green,thick] ([shift={(0.01, 0.01)}]\corls, \corbs) rectangle ([shift={(-0.01, -0.01)}]{\corls+\corrts}, {\corbs+\corrtys});
    \end{scope}
    \end{tikzpicture}} &
    
    \multicolumn{2}{c}{
    \begin{tikzpicture}
    \node[anchor=south west,inner sep=0] (image) at (0,0) {\includegraphics[trim=0 500px 900px 100px, clip,  width=\wb\columnwidth]{data/experiments/effect/figures/Ours/\img}};
    \begin{scope}[x={(image.south east)},y={(image.north west)}]
        \draw[red,thick] ([shift={(0.01, 0.01)}]\corl, \corb) rectangle ([shift={(-0.01, -0.01)}]{\corl+\corrt}, {\corb+\corrty});
        \draw[green,thick] ([shift={(0.01, 0.01)}]\corls, \corbs) rectangle ([shift={(-0.01, -0.01)}]{\corls+\corrts}, {\corbs+\corrtys});
    \end{scope}
    \end{tikzpicture}} &
    
    \multicolumn{2}{c}{
    \begin{tikzpicture}
    \node[anchor=south west,inner sep=0] (image) at (0,0) {\includegraphics[trim=0 500px 900px 100px, clip,  width=\wb\columnwidth]{data/DPDD/figures/GT/\img}};
    \begin{scope}[x={(image.south east)},y={(image.north west)}]
        \draw[red,thick] ([shift={(0.01, 0.01)}]\corl, \corb) rectangle ([shift={(-0.01, -0.01)}]{\corl+\corrt}, {\corb+\corrty});
        \draw[green,thick] ([shift={(0.01, 0.01)}]\corls, \corbs) rectangle ([shift={(-0.01, -0.01)}]{\corls+\corrts}, {\corbs+\corrtys});
    \end{scope}
    \end{tikzpicture}}
    
    \\[-0.03in]
    
    \adjustbox{width=\ws\columnwidth, trim={\corl\width} {\corb\height} {\width-(\corl\width+\corrt\width)} {\height-(\corb\height+\corrty\height)},clip,cfbox=red 0.7pt -0.7pt}
    {\includegraphics[trim=0 500px 900px 100px, clip,  width=\ws\columnwidth]{data/DPDD/figures/Input/\img}} &
    \adjustbox{width=\ws\columnwidth, trim={\corls\width} {\corbs\height} {\width-(\corls\width+\corrts\width)} {\height-(\corbs\height+\corrtys\height)}, clip,cfbox=green 0.7pt -0.7pt}
    {\includegraphics[trim=0 500px 900px 100px, clip,  width=\ws\columnwidth]{data/DPDD/figures/Input/\img}} &

    \adjustbox{width=\ws\columnwidth, trim={\corl\width} {\corb\height} {\width-(\corl\width+\corrt\width)} {\height-(\corb\height+\corrty\height)},clip,cfbox=red 0.7pt -0.7pt}
    {\includegraphics[trim=0 500px 900px 100px, clip,  width=\ws\columnwidth]{data/experiments/effect/figures/FAC/\img}} &
    \adjustbox{width=\ws\columnwidth, trim={\corls\width} {\corbs\height} {\width-(\corls\width+\corrts\width)} {\height-(\corbs\height+\corrtys\height)}, clip,cfbox=green 0.7pt -0.7pt}
    {\includegraphics[trim=0 500px 900px 100px, clip,  width=\ws\columnwidth]{data/experiments/effect/figures/FAC/\img}} &
    
    \adjustbox{width=\ws\columnwidth, trim={\corl\width} {\corb\height} {\width-(\corl\width+\corrt\width)} {\height-(\corb\height+\corrty\height)},clip,cfbox=red 0.7pt -0.7pt}
    {\includegraphics[trim=0 500px 900px 100px, clip,  width=\ws\columnwidth]{data/experiments/effect/figures/Ours/\img}} &
    \adjustbox{width=\ws\columnwidth, trim={\corls\width} {\corbs\height} {\width-(\corls\width+\corrts\width)} {\height-(\corbs\height+\corrtys\height)}, clip,cfbox=green 0.7pt -0.7pt}
    {\includegraphics[trim=0 500px 900px 100px, clip,  width=\ws\columnwidth]{data/experiments/effect/figures/Ours/\img}} &

    \adjustbox{width=\ws\columnwidth, trim={\corl\width} {\corb\height} {\width-(\corl\width+\corrt\width)} {\height-(\corb\height+\corrty\height)}, clip,cfbox=red 0.7pt -0.7pt}
    {\includegraphics[trim=0 500px 900px 100px, clip,  width=\ws\columnwidth]{data/DPDD/figures/GT/\img}} &
    \adjustbox{width=\ws\columnwidth, trim={\corls\width} {\corbs\height} {\width-(\corls\width+\corrts\width)} {\height-(\corbs\height+\corrtys\height)}, clip,cfbox=green 0.7pt -0.7pt}
    {\includegraphics[trim=0 500px 900px 100px, clip,  width=\ws\columnwidth]{data/DPDD/figures/GT/\img}}
    
    \\[-0.02in]
    \multicolumn{2}{c}{(a) Input} &
    \multicolumn{2}{c}{(b) FAC} &
    \multicolumn{2}{c}{(c) IAC} &
    \multicolumn{2}{c}{(d) GT}

  \end{tabular}

 \caption{Qualitative comparison between the FAC and IAC layers evaluated on the DPDD dataset \cite{Abuolaim:2020:DPDNet}.
FAC in (b) means our final model whose IAC layers are replaced with FAC layers.
IAC in (c) means our final model with IAC layers.
The input blurred image has large defocus blur, so details in the red and green boxes are not visible.
Our final model with IAC shows better restored details compared to the model with FAC.
}
  \vspace{-12pt}
  \label{fig:effect:IAC}
\end{figure}

\section{Conclusion}
This paper proposed a single image defocus deblurring framework based on our novel Iterative Filter Adaptive Network (IFAN).
IFAN flexibly handles spatially-varying defocus blur by predicting per-pixel separable deblurring filters.
For efficient and effective handling of a large blur, we proposed Iterative Adaptive Convolution (IAC) that iteratively applies separable filters on features.
In addition, IFAN learns to estimate defocus blur from a single image more accurately through the learning of disparity map estimation and reblurring.
In the experiments, we verified the effect of each component in our model, and showed that our method achieves state-of-the-art performance.

The proposed network is still limited in handling significantly large defocus blur (\eg, \Fig{\ref{fig:effect:IAC}}c has remaining blur).
Our network works best with a typical isotropic defocus blur, 
but may not properly handle blur with irregular shape (\eg, swirly bokeh) or strong highlight (\ie, glitter bokeh).
Refer to the supplementary material for the failure cases.
We plan to extend our RealDOF test set by adding more image pairs of diverse blur types captured with various cameras and lenses.

\vspace{-13pt}
\small
\paragraph{\small Acknowledgements}
This work was supported by the Ministry of Science and ICT, Korea,
through 
IITP grants
(SW Star Lab, 2015-0-00174;
Artificial Intelligence Graduate School Program (POSTECH), 2019-0-01906)
and
NRF grants (2018R1A5A1060031; 2020R1C1C1014863).

{\small
\bibliographystyle{compile/ieee_fullname}
\bibliography{ms}
}
\end{document}